# Experimental Comparison of Representation Methods and Distance Measures for Time Series Data

Xiaoyue Wang · Hui Ding · Goce Trajcevski ·
Peter Scheuermann · Eamonn Keogh

This paper is a draft of a paper under review at DMKD journal.

If you have any comments, please send them to the authors.

If you like, you can CC the editor Geoff Webb (Geoff.Webb@monash.edu),

who will past them onto the action editor and reviewers.

**Abstract** The previous decade has brought a remarkable increase of the interest in applications that deal with querying and mining of time series data. Many of the research efforts in this context have focused on introducing new representation methods for dimensionality reduction or novel similarity measures for the underlying data. In the vast majority of cases, each individual work introducing a particular method has made specific claims and, aside from the occasional theoretical justifications, provided quantitative experimental observations. However, for the most part, the comparative aspects of these experiments were too narrowly focused on demonstrating the benefits of the proposed methods over some of the previously introduced ones. In order to provide a comprehensive validation, we conducted an extensive experimental study re-

Xiaoyue Wang
University of California Riverside
E-mail: xwang@cs.ucr.edu

Hui Ding
Northwestern University
E-mail: hdi117@eecs.northwestern.edu

Goce Trajcevski
Northwestern University
E-mail: goce@eecs.northwestern.edu

Peter Scheuermann
Northwestern University
E-mail: peters@eecs.northwestern.edu

Eamonn Keogh
University of California Riverside
E-mail: eamonn@cs.ucr.edu

implementing eight different time series representations and nine similarity measures and their variants, and testing their effectiveness on thirty-eight time series data sets from a wide variety of application domains. In this paper, we give an overview of these different techniques and present our comparative experimental findings regarding their effectiveness. In addition to providing a unified validation of some of the existing achievements, our experiments also indicate that, in some cases, certain claims in the literature may be unduly optimistic.

## 1 Introduction

Time series data are being generated at an unprecedented scale and rate from almost every application domain, e.g., daily fluctuations of stock market, traces of dynamic processes and scientific experiments, medical and biological experimental observations, various readings obtained from sensor networks, position updates of moving objects in location-based services, etc. As a consequence, in the last decade there has been a dramatically increasing amount of interest in querying and mining such data which, in turn, resulted in a large amount of work introducing new methodologies for indexing, classification, clustering and approximation of time series [18, 24, 31].

The main goals of managing time series data are the *effectiveness* and the *efficiency*, and the two key aspects towards achieving them are: (1) *representation methods*, and (2) *similarity measures*. Time series are essentially *high dimensional* data [24] and working directly with such data in its raw format is very expensive in terms of both processing and storage cost. It is thus highly desirable to develop representation techniques that can reduce the dimensionality of time series, while still preserving the fundamental characteristics of a particular data set. In addition, unlike canonical data types, e.g., nominal/categorical or ordinal variables [46], where the distance definition between two values is usually fairly straightforward, the *distance* between time series needs to be carefully defined in order to properly capture the semantics and reflect the underlying (dis)similarity of such data. This is particularly desirable for similarity-based retrieval, classification, clustering and other querying and mining tasks over time series data [24].

Many techniques have been proposed for representing time series with reduced dimensionality, for example: *Discrete Fourier Transformation* (DFT) [18], *Single Value Decomposition* (SVD) [18], *Discrete Cosine Transformation* (DCT) [39], *Discrete Wavelet Transformation* (DWT) [51], *Piecewise Aggregate Approximation* (PAA) [33], *Adaptive Piecewise Constant Approximation* (APCA) [32], *Chebyshev polynomials* (CHEB) [11], *Symbolic Aggregate approXimation* (SAX) [43], *Indexable Piecewise Linear Approximation* (IPLA) [16], etc. In conjunction with these techniques, there are over a dozen distance measures used for evaluating similarity of time series presented in the literature, e.g., *Euclidean distance* (ED) [18], *Dynamic Time Warping* (DTW) [10,35], distance based on *Longest Common Subsequence* (LCSS) [61], *Edit Distance with Real Penalty* (ERP) [13], *Edit Distance on Real sequence* (EDR) [14], DISSIM [20], *Sequence Weighted Alignment model* (Swale) [45], *Spatial Assembling Distance* (SpADe) [17] and similarity search based on *Threshold Queries* (TQuEST) [7]. Quite a few of these works, as well as some of their extensions, have been widely cited in the literature and applied to facilitate query processing and data mining of time series data.

Given the multitude of competitive techniques, we believe that there is a strong need for a comprehensive comparison which, in addition to providing a foundation for benchmarks, may also reveal certain omissions in the comparative observations reported in the individual works. In the common case, every newly-introduced representation method or distance measure claimed a particular superiority over some of the existing results. However, it has been demonstrated that some empirical evaluations have been inadequate [34] and, worse yet, some of the claims are even contradictory. For example, one paper claims *"wavelets outperform the DFT"* [52], another claims *"DFT filtering performance is superior to DWT"* [28] and yet another claims *"DFT-based and DWT-based techniques yield comparable results"* [63]. Clearly, not all of these claims can be simultaneously true. An important consequence of this observation is that there is risk that such (or similar) claims may not only cause a confusion to newcomers and practitioners in the field, but also cause a waste of time and research efforts due to assumptions based on incomplete or incorrect claims.

Motivated by these observations, we have conducted the most extensive set of time series experiments to-date, re-evaluating the state-of-the-art representation methods and similarity measures for time series that appeared in high quality conferences and journals. Specifically, as the main contributions of this work, we have:

- Re-implemented 8 different representation methods for time series, and compared their *pruning power* over various time series data sets.
- Re-implemented 9 different similarity measures and their variants, and compared their effectiveness using 38 real world data sets from highly diverse application domains.
- Provided certain analysis and conclusions based on the experimental observations.

We note that all of our source code implementations and the data sets are publicly available on our website [1].

The rest of this paper is organized as follows. Section 2 reviews the concept of time series, and gives an overview of the definitions of different representation techniques and similarity measures investigated in this work. Section 3 and Section 4 present the main contribution of this work – the results of the extensive experimental evaluations of different representation methods and similarity measures, respectively. In Section 5, we summarize some of the myths and misunderstandings about DTW. Section 6 concludes the paper and discusses possible future extensions of the work.

## 2 Preliminaries

Typically, most of the existing works on time series assume that *time* is discrete. For simplicity and without any loss of generality, we make the same assumption here. Formally, a *time series* data is defined as a sequence of pairs $T = [(p_1, t_1), (p_2, t_2), ..., (p_i, t_i), ..., (p_n, t_n)]$ $(t_1 < t_2 < ... < t_i < ... < t_n)$, where each $p_i$ is a data point in a $d$-dimensional data space, and each $t_i$ is the time stamp at which the corresponding $p_i$ occurs[1]. If the sampling rates of two time series are the same, one can omit the time stamps and consider them as sequences of $d$-dimensional data points. Such a sequence is called the *raw representation* of the time series. In reality however, sampling rates of time series may be different. Furthermore, some data points of time series may be dampened by noise or even completely missing, which poses additional challenges to the processing of such data. For a given time series, its number of data points $n$ is called its *length*. The portion of a time series between two points $p_i$ and $p_j$ (inclusive) is called a *segment* and is denoted as $s_{ij}$. In particular, the segment $s_{i(i+1)}$ between two consecutive points is called a *line segment*.

In the following subsections, we briefly review the representation methods and similarity measures studied in this work. We note that this is not intended to be a complete survey of the available techniques and is only intended to provide the necessary background for following and understanding our experimental evaluations.

## 2.1 Representation Methods for Time Series

There is a plethora of time series representation methods, each of them proposed for the purpose of supporting similarity search and data mining tasks.

A classification of the major techniques, organized in a hierarchical manner, is shown in Figure 1.

As illustrated, there are two basic categories:

- *Data Adaptive* representations: in this category, a common representation will be chosen for all items in the database that minimizes the global reconstruction error.
- *Non-Data Adaptive* representations: in contrast, these methods consider local properties of the data, and construct an approximate representation accordingly.

For example, Adaptive Piecewise Constant Approximation (APCA, an adaptive technique) transforms each time series by a set of constant value segments of varying lengths such that their individual reconstruction errors are minimal. On the other hand, Piecewise Aggregate Approximation(PAA, a non-adaptive technique), approximates a time series by dividing it into equal-length segments and recording the mean value of the datapoints that fall within the segment.

The representations annotated with an asterisk (∗) in Figure 1 have the very desirable property of allowing *lower bounding*. This property, essentially, allows one to define a distance measure that can be applied to the reduced-size (i.e., compressed) representations of the corresponding time series, that is guaranteed to be less than or equal to the true distance which is measured on the raw data. The main benefit of the lower bounding property is that it allows using the respective reduced-size representations to index the data, with a guarantee of *no false negatives* [18]. The list of representations considered in this study includes (in approximate order of introduction) DFT, DCT, DWT, PAA, APCA, SAX, CHEB and IPLA. The only lower bounding omissions from our experiments below are the eigenvalue analysis techniques such as SVD and PCA [39]. While such techniques give optimal linear dimensionality reduction, we believe they are untenable for large data sets. For example, while [59] notes that they can transform 70000 time series in under 10 minutes, the assumption is that the data is memory resident. However, transforming out-of-core (disk resident) data sets using these methods becomes unfeasible. Note that the available literature seems to agree with us on this point. For (at least) DFT, DWT and PAA, there are more than a dozen projects that use these representations to index over 100000 objects for query-by-humming [27,70], Mo-Cap indexing [12], etc. At the time of writing this article, however, we are unaware of any projects of a similar scale that use SVD.

---

[1] We do not differentiate between the *time of occurrence* and the *time of detection* of a particular event in this work [9] or, to phrase it in a different context – we do not distinguish the *valid time* from the *transaction time* [60]

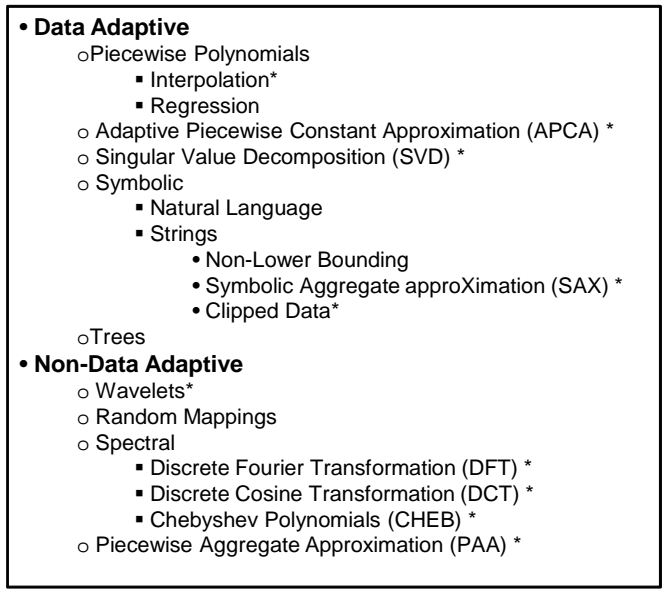

**Fig. 1:** A Hierarchy of Representation Methods

## 2.2 Similarity Measures for Time Series

We now give an overview of the 9 similarity measures evaluated in this work which, for convenience, are summarized in Figure 2.

Given two time series $T_1$ and $T_2$, a similarity function $Dist$ calculates the distance between the two time series, denoted by $Dist(T_1, T_2)$. In the following we will refer to distance measures that compare the $i-$th point of one time series to the $i-$th point of another as *lock-step measures* (e.g., Euclidean distance and the other Lp norms), and distance measures that allow comparison of one-to-many points (e.g., DTW) and one-to-many/one-to-none points (e.g., LCSS) as *elastic measures*. Figures 3 through 6 provide illustrations of the corresponding intuitions behind the major classes of distance measures. Note that in every case, the two time series are shown shifted apart in the y-axis for visual clarity, however they would typically be normalized and therefore overlapping [34]. Figure 3 shows the intuition behind Lock Step measures, a class which includes the ubiquitous Euclidean distance.

- **Lock-step Measure**
  - $L_p$-norms
    - $L_1$-norm (Manhattan Distance)
    - $L_2$-norm (Euclidean Distance)
    - $L_{inf}$-norm
  - DISSIM
- **Elastic Measure**
  - Dynamic Time Warping (DTW)
  - Edit distance based measure
    - Longest Common SubSequence (LCSS)
    - Edit Sequence on Real Sequence (EDR)
    - Swale
    - Edit Distance with Real Penalty (ERP)
- **Threshold-based Measure**
  - Threshold query based similarity search (TQuEST)
- **Pattern-based Measure**
  - Spatial Assembling Distance (SpADe)

**Fig. 2:** A Summary of Similarity Measures

The most straightforward similarity measure for time series is the *Euclidean Distance* [18] and its variants, based on the common $L_p$-norms [65]. In particular, in this work we used $L_1$ (Manhattan), $L_2$ (Euclidean) and $L_\infty$ (Maximum) norms (c.f. [65]). In the sequel, the terms Euclidean distance and $L_2$ norm will be used interchangeably. In addition to being relatively straightforward for intuitive understanding, the Euclidean distance and its variants have several other advantages. An important one is that the complexity of evaluating these measures is linear, and they are easy to implement and indexable with any access method and, in addition, they are *parameter-free*. Furthermore, as we will demonstrate, the Euclidean distance is surprisingly competitive with the other, more complex approaches, especially if the size of the training set/database is relatively large. However, since the mapping between the points of two time series is *fixed*, these distance measures are very sensitive to noise and misalignments in time, and are unable to handle *local time shifting*, i.e., similar segments that are out of phase.

The DISSIM distance [20] aims at computing the similarity of time series with different sampling rates. However, the original similarity function is numerically too difficult to compute, and the authors proposed an approximated distance with a formula for computing the error bound.

Inspired by the need to handle time warping in similarity computation, Berndt and Clifford [10] introduced DTW, a classical speech recognition tool, to the data mining community, in order to allow a time series to be "stretched" or "compressed" to provide a better match with another time series. Figure 4 illustrates the intuition behind DTW and other elastic measures.

Several lower bounding measures have been introduced to speed up similarity search using DTW [30,35,37,66], and it has been shown that the amortized cost for computing DTW on large data sets is linear [30,35]. The original DTW distance is also parameter free, however, as has been reported in [35,62] enforcing a *temporal constraint* $\delta$ on the warping window size of DTW not only improves its computation efficiency, but also improves its accuracy for measuring time series similarity, as extended warping may introduce pathological matchings between two time series and distort the true similarity. The constraint warping is also utilized for developing the lower-bounding distance [35] as well as for indexing time series based on DTW [62].

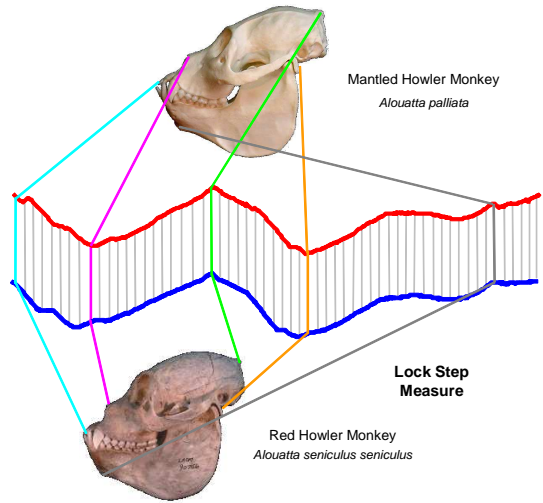

**Fig. 3:** An illustration of a *Lock Step* measure. Note the "one-to-one" mapping of data points. The distance measure is proportional to the length of the gray lines.

Another group of similarity measures for time series has been developed based on the concept of the *edit distance* for strings. The main intuition behind the Editing measures is visualized in Figure 5.

The best known example from this category is the LCSS distance, which is based on the *longest common subsequence* model [5,61]. To adapt the concepts used in matching characters and strings in the settings of time series, a *threshold parameter $\varepsilon$* was introduced, the semantics of which is that two points from two time series are considered to match if their distance is less than $\varepsilon$. The work reported in [61] also took into consideration an additional constraint – the matching of points along the temporal dimension, using a so called *warping threshold $\delta$*. A lower-bounding measure and indexing technique for LCSS were introduced in [62].

EDR [14] is another similarity measure based on the edit distance. Similar to LCSS, EDR also uses a threshold parameter $\varepsilon$, except its role is to quantify the distance between a pair of points to 0 or 1. Unlike LCSS, EDR assigns penalties to the gaps between two matched segments according to the lengths of the gaps.

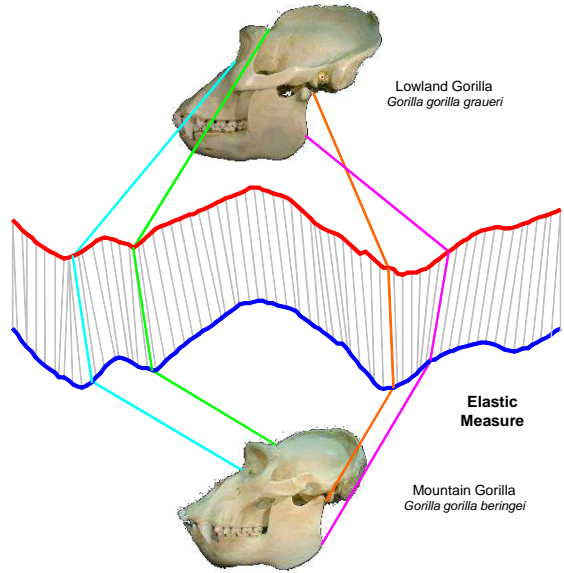

**Fig. 4:** An illustration of *Elastic* measure. Note that unlike Lock Step measures, here we allow the possibility of "one-to-many" mapping of the data points, but each data point must be matched. The distance measure is proportional to the length of the gray lines.

The ERP distance [13] attempts to combine the merits of both DTW and EDR, by introducing the concept of a *constant reference point* for computing the distance between gaps of two time series. Essentially, if the distance between two points is too large, ERP simply uses the distance value between one of those point and the reference point.

Recently, a new approach for computing the edit distance based similarity measures was proposed in [45]. Whereas traditional tabular dynamic programming was used for computing DTW, LCSS, EDR and ERP, a matching threshold is used to divide the data space into grid cells and, subsequently, matching points are found by hashing. The similarity model Swale is proposed that rewards matching points and penalizes gaps. In addition to the matching threshold $\varepsilon$, Swale requires the tuning of two parameters: the *matching reward* weight $r$ and the *gap penalty* weight $p$.

The TQuEST distance [7] introduced a rather novel approach to computing the similarity measure between time series. The main idea behind TQuEST is that, given a threshold parameter $\tau$, a time series is transformed into a sequence of so-called *threshold-crossing* time intervals, where the points within each time interval have a value greater than $\tau$. Each time interval is then treated as a point in a two dimensional space, where the starting time and ending time constitute the two dimensions. The similarity between two time series is then defined as the Minkowski sum of the two sequences of time interval points [19]. Figure 6 visually illustrates the intuition behind the threshold measures.

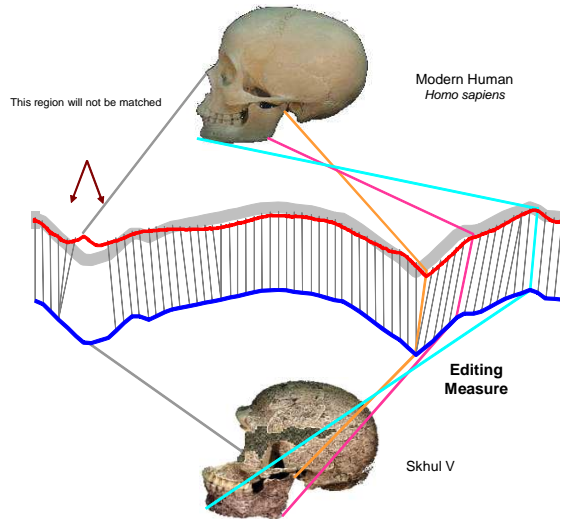

**Fig. 5:** An illustration of an *Editing* measure. Note that, similarly to the Elastic measures, we allow the possibility of "one-to-many" mapping of the data points. However in addition, we also allow the possibility of *not* matching some (one or more) points. The distance measure is proportional to the length of the gray lines.

The last approach considered in this work is SpADe [17], which is a pattern-based similarity measure for time series. The key idea behind the presented algorithm is to find out matching segments within the entire time series, called *patterns*, by allowing shifting and scaling in both the temporal and amplitude dimensions. The problem of computing similarity value between time series is then transformed to the one of finding the most similar set of matching patterns. A peculiarity of SpADe is that it requires tuning a number of parameters, such as the temporal scale factor, amplitude scale factor, pattern length, sliding step size,etc.

## 3 Comparison of Time Series Representations

We compare all the major time series representations that have been proposed in the literature, including SAX, DFT, DWT, DCT, PAA, CHEB, APCA and IPLA. We note that all the representation methods studied in this paper allow lower bounding, and any of them can be used to index the Euclidean Distance, the Dynamic Time

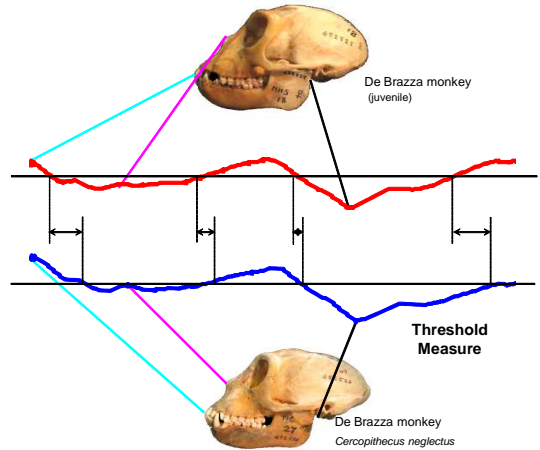

**Fig. 6:** An illustration of a *Threshold* measure. The distance measure is proportional to the length of the double-headed arrow.

Warping, and at least some of the other elastic measures. While various subsets of these representations have been compared before, to the best of our knowledge, this is the first attempt to compare all of them together. One obvious question that needs to be considered is what metric should should be used for comparison? We postulate that the wall clock time is a poor choice, because it may be open to an implementation bias [34]. Instead, we believe that using the *tightness of lower bounds* (TLB) is a very meaningful measure [33], and this also appears to be the current consensus in the literature [11, 13, 16, 30–32, 35, 53, 62]. Formally, given two time series, $T$ and $S$, the corresponding TLB is defined as

$$TLB = LowerBoundDist(T, S)/TrueEuclideanDist(T, S)$$

The advantage of using TLB is two-fold:

1. It is a completely *implementation-free* measure, independent of hardware and software choices, and is therefore completely reproducible.
2. It allows a *very accurate prediction* of the indexing performance.

If the value of TLB is zero, then any indexing technique is condemned to retrieving every single time series from the disk. On the other hand, if the value of TLB is one then, after some trivial processing in main memory, we could simply retrieve a single object from the disk and guarantee that we have obtained the true nearest neighbor.

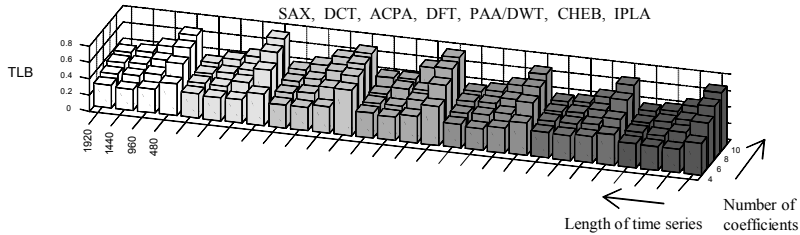

**Fig. 7:** The tightness of lower bounds(TLB) for various time series representations on an ECG data set.

Note that, in general, the speedup obtained is non-linear in TLB, that is to say, if one representation has a lower bound that is twice as large as another, we can usually expect a much greater than two-fold decrease in the number of disk accesses.

As part of this work, we randomly sampled $T$ and $S$ (with replacement) 1000 times for each combination of parameters. We varied the time series length among the values of $\{480, 960, 1440, 1920\}$, as well as the number of coefficients per time series available to the dimensionality reduction approach among the values of $\{4, 6, 8, 10\}$ (each coefficient takes 4 bytes). For SAX, we hard coded the cardinality to 256. Figure 7 shows the result of one such experiment with an ECG data set.

At a first glance, the results of this experiment may appear surprising, as they show that there is very little difference between representations, in spite of the claims to the contrary in the literature. However, we believe that most of these claims may be due to some errors or bias in the experiments. For example, it was recently claimed that DFT is much worse than all the other approaches [16], however it appears that the complex conjugate property of DFT was not exploited. As another example, it was claimed *"it only takes* 4 *to* 6 *Chebyshev coefficients to deliver the same pruning power produced by* 20 *APCA coefficients"* [11], however this claim has since been withdrawn by the authors [2]. Of course there are some variabilities and differences depending on the data sets. For example, on a highly periodic data set the spectral methods are better, and on bursty data sets APCA can be significantly better, as shown in Figure 8.

In contrast, in Figure 9 we can see that highly periodic data can slightly favor the spectral representations (DCT, DFT, CHEB) over the polynomial representations (SAX, APCA, DWT/PAA, IPLA).

However it is worth noting that the differences presented in these figures are the most extreme cases found in a search spanning over 80 diverse data sets from the publicly available UCR Time Series Data Mining Archive [29]. This, in turn, makes it very likely that, in general, there is very little to choose between representations in terms of pruning power.

## 4 Comparison of Time Series Similarity Measures

In this section we present our experimental evaluation on the accuracy of different similarity measures.

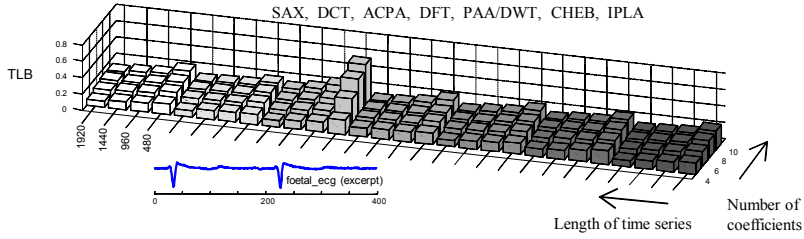

**Fig. 8:** The tightness of lower bounds(TLB) for various time series representations on a relatively bursty data set (see inset).

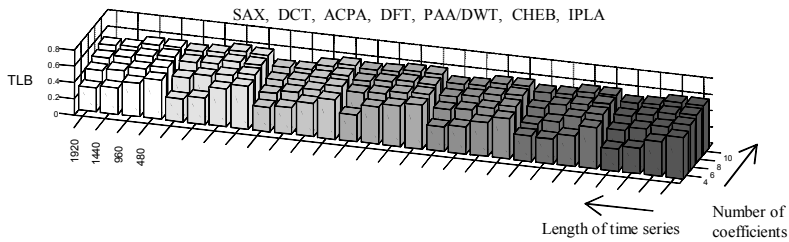

**Fig. 9:** The tightness of lower bounds(TLB) for various time series representations on a periodic data set of tide levels.

## 4.1 The Effect of Data Set Size on Accuracy and Speed

We first discuss an extremely important finding which, in some circumstances makes some of the previous findings on efficiency, and the subsequent findings on accuracy, moot. This finding has been noted before [53], but does not seem to be appreciated by the database community.

For an elastic distance measure, both the *accuracy* of classification (or precision/recall of similarity search), and the *amortized speed*, depend critically on the size of the data set. Specifically, on one hand, as data sets get larger, the amortized speed of elastic measures approaches that of lock-step measures, on the other hand, the accuracy/precision of lock-step measures approaches that of the elastic measures. This observation has significant implications for much of the research in the literature. Many papers claim something like *"I have shown on these* 80 *time series that my elastic approach is faster than DTW and more accurate that Euclidean distance, so if you want to index a million time series, use my method"*. However our observation suggests that even if the method is faster than DTW, the speed difference will decrease for larger data sets. Furthermore, for large data sets, the differences in accuracy/precision will also diminish or disappear. To demonstrate our claim we conducted experiments on two highly warped data sets that are often used to highlight the superiority of elastic measures, Two-Patterns and CBF. Because these are synthetic data sets, one has the luxury of creating as many instances as needed, using the data generation algorithms proposed in the original papers [21, 22]. However, it is critical to note that the same effect can be seen on all the data sets considered in this work. For each problem we created 10000 test time series, and increasingly large training data sets of size 50, 100, 200, ..., 6400. We measured the classification accuracy of 1NN for the various data sets (explained in more detail in Section 4.2.1), using both Euclidean distance and DTW with 10% warping window, and the results are shown in Figure 10.

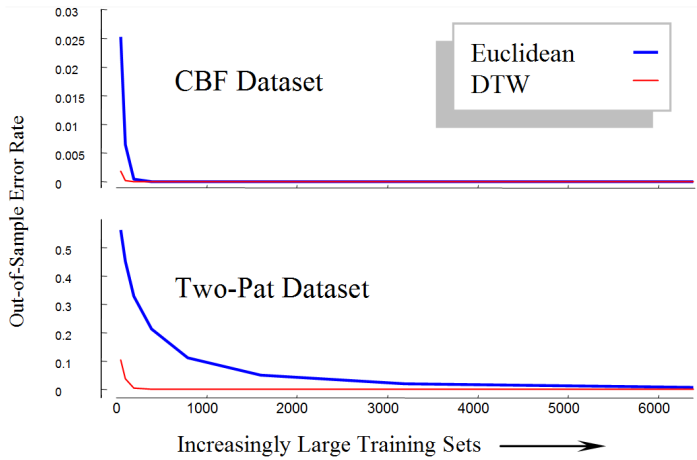

**Fig. 10:** The error rate for 1-Nearest Neighbor Classification for increasingly large instantiations of two classic time series benchmarks.

Note that for small data sets, DTW is significantly more accurate than Euclidean distance in both cases. However, for CBF, by the time we have a mere 400 time series in our training set, there is no statistically significant difference. For Two-Patterns it takes longer for Euclidean Distance to converge to DTW's accuracy, nevertheless, by the time we have seen a few thousand objects there is no statistically significant difference.

This experiment can also be used to demonstrate our claim that the amortized speed of a (lower-boundable) elastic method approaches that of Euclidean distance. Recall that Euclidean distance has a time complexity of $O(n)$ and that a single DTW calculation has a time complexity of $O(nw)$, where $w$ is the warping window size. However for similarity search or 1NN classification, the amortized complexity of DTW is $O((P \cdot n) + (1 - P) \cdot nw)$, where $P$ is the fraction of DTW calculations pruned by a linear time lower bound such as LB_Keogh. A similar result can be achieved for LCSS as well. In the Two-Pattern experiments above, when classifying with only 50 objects, $P = 0.1$, so we are forced to do many full DTW calculations. However, by the time we have 6400 objects, we empirically find out that $P = 0.9696$, so about 97% of the objects are disposed of in the same time as it takes to do a Euclidean distance calculation. To ground this into concrete numbers, it takes less that one second to find the nearest neighbor to a query in the database of 6400 Two-Patterns time series, on our off-the-shelf desktop, even if we use the pessimistically wide warping window. We note that this time is for just sequential search with a lower bound – no attempt was made to index the data.

To summarize, many of the claims over who has the fastest or most accurate distance measure have been biased by the lack of tests on very (or even slightly) large data sets.

## 4.2 Accuracy of Similarity Measures

In this section, we evaluate the accuracy of the similarity measures presented in Section 2. We first explain the methodology of our evaluation, as well as the parameters that need to be tuned for each similarity measure. We then present the results of our experiments and discuss several interesting findings.

### 4.2.1 Accuracy Evaluation Framework

Accuracy evaluation answers one of the most important questions about a similarity measure: why is this a good measure for describing the (dis)similarity between time series? Surprisingly, we found that the accuracy evaluation is often insufficient in existing literature: it has been either based on subjective evaluation, e.g., [7,14], or using clustering with small data sets which are not statistically significant, e.g., [45,62]. In this work, we use an objective evaluation method recently proposed [34]. The idea is to use a one nearest neighbor (1NN) classifier [24,47] on labelled data to evaluate the efficacy of the distance measure used. Specifically, each time series has a correct class label, and the classifier tries to predict the label as that of its nearest neighbor in the training set. There are several advantages with this approach. First, it is well known that the underlying distance metric is critical to the performance of 1NN classifier [47], hence, the accuracy of the 1NN classifier directly reflects the effectiveness of the similarity measure. Second, the 1NN classifier is straightforward to implement and is parameter free, which makes it easy for anyone to reproduce our results. Third, it has been proved that the error ratio of 1NN classifier is at most twice the Bayes error ratio [55]. Finally, we note that while there have been attempts to classify time series with decision trees, neural networks, Bayesian networks, supporting vector machines, etc., the best published results (by a large margin) come from simple nearest neighbor methods [64].

To evaluate the effectiveness of each similarity measure, we use a cross-validation algorithm as described in Algorithm 1, based on the approach suggested in [57]. We first use a stratified random split to divide the input data set into $k$ subsets for the subsequent classification (line 1) in order to minimize the impact of skewed class distribution. The number of cross validations $k$ is dependent on the data sets and we explain shortly how we choose the proper value for $k$. We then carry out the cross validation, using one subset at a time for the training set of the 1NN classifier, and the rest $k-1$ subsets as the testing set (lines $3-9$). If the similarity measure $SimDist$ requires parameter tuning, we divide the training set into two equal size stratified subsets, and use one of the subset for parameter tuning (lines $4-7$). We perform an exhaustive search for all the possible (combinations of) value(s) of the similarity parameter, and conduct a leave-one-out classification test with a 1NN classifier. We record the error ratios of the leave-one-out test, and use the parameter values that yield the minimum

---

**Algorithm 1** Time Series Classification with 1NN Classifier

---

**Input:** Labelled time series data set $\mathbb{T}$, similarity measure operator $SimDist$, number of crosses $k$
**Output:** Average 1NN classification error ratio and standard deviation
 1: Randomly divide $\mathbb{T}$ into $k$ stratified subsets $\mathbb{T}_1, \ldots, \mathbb{T}_k$
 2: Initialize an array $ratios[k]$
 3: **for** Each subset $\mathbb{T}_i$ of $\mathbb{T}$ **do**
 4:     **if** $SimDist$ requires parameter tuning **then**
 5:         Randomly split $\mathbb{T}_i$ into two equal size stratified subsets $\mathbb{T}_{i1}$ and $\mathbb{T}_{i2}$
 6:         Use $\mathbb{T}_{i1}$ for parameter tuning, by performing a leave-one-out classification with 1NN classifier
 7:         Set the parameters to values that yields the minimum error ratio from the leave-one-out tuning process
 8:     Use $\mathbb{T}_i$ as the training set, $\mathbb{T} - \mathbb{T}_i$ as the testing set
 9:     $ratio[i] \leftarrow$ the classification error ratio with 1NN classifier
10: **return** Average and standard deviation of $ratios[k]$

---

error ratio. Finally, we report the average error ratio of the 1NN classification over the $k$ cross validations, as well as the standard deviation (line 10).

Algorithm 1 requires that we provide an input $k$ for the number of cross validations. In our experiments, we need to take into consideration the impact of training data set size discussed in Section 4.1. Therefore, our selection of $k$ for each data set attempts to strike a balance between the following factors:

1. The training set size should be selected to enable discriminativity, i.e., one can tell the performance difference between different distance measures.

2. The number of items in the training set should be large enough to represent each class. This is especially important when the distance measure needs parameter tuning.

3. The number of cross validations should be between $5 - 20$ in order to minimize bias and variation, as recommended in [38].

The actual number of splits is empirically selected such that the training error for 1NN Euclidean distance (which we use as a comparison reference) is not perfect, but significantly better than the default rate.

Several of the similarity measures that we investigated require the setting of one or more parameters. The proper values for these parameters are key to the effectiveness of the measure. However, most of the time only empirical values are provided for each parameter in isolation. In our experiments, we perform an exhaustive search for all the possible values of the parameters, as described in Table 1.

For DTW and LCSS measures, a common optional parameter is the window size $\delta$ that constrains the temporal warping, as suggested in [62]. In our experiments we consider both the version of distance measures without warping and with warping. For the latter case, we search for the best warping window size up to 25% of the length of the time series $n$. An additional parameter for LCSS, which is also used in EDR and Swale, is the matching threshold $\varepsilon$. We search for the optimal threshold starting from $0.02 \cdot Stdv$ up to $Stdv$, where $Stdv$ is the standard deviation of the data set. Swale has two other parameters, the matching reward weight and the gap penalty weight. We fix the matching reward weight to 50 and search for the optimal penalty weight from 0 to 50, as suggested by the authors. Although the warping window size can also be constrained for EDR, ERP and Swale, we only consider full matching for these distance measures in our current experiments – and the rationale for this choice was the fairness. Namely, while each of the three approaches may be amenable to less-than-full matching, this was never proposed, nor considered, as a feature in the original works. For TQuEST, we search for the optimal querying threshold from $Avg - Stdv$ to $Avg + Stdv$, where $Avg$ is the average of the time series data set. For SpADe, we tune four parameters based on the original implementation and use the parameter tuning strategy, i.e. search range, step size, as suggested by the authors. In Table 1, *plength* is the length of the patterns, *ascale* and *tscale* are the maximum amplitude and temporal scale differences allowed respectively, and *slidestep* is the minimum temporal difference between two patterns.

| Parameter | Min Value | Max Value | Step Size |
|---|---|---|---|
| DTW.$\delta$ | 1 | $25\% \cdot n$ | 1 |
| LCSS.$\delta$ | 1 | $25\% \cdot n$ | 1 |
| LCSS.$\varepsilon$ | $0.02 \cdot Stdv$ | $Stdv$ | $0.02 \cdot Stdv$ |
| EDR.$\varepsilon$ | $0.02 \cdot Stdv$ | $Stdv$ | $0.02 \cdot Stdv$ |
| Swale.$\varepsilon$ | $0.02 \cdot Stdv$ | $Stdv$ | $0.02 \cdot Stdv$ |
| Swale.$reward$ | 50 | 50 | - |
| Swale.$penalty$ | 0 | $reward$ | 1 |
| TQuEST.$\tau$ | $Avg - Stdv$ | $Avg + Stdv$ | $0.02 \cdot Stdv$ |
| SpADe.$plength$ | 8 | 64 | 8 |
| SpADe.$ascale$ | 0 | 4 | 1 |
| SpADe.$tscale$ | 0 | 4 | 1 |
| SpADe.$slidestep$ | $plength/32$ | $plength/8$ | $plength/32$ |

**Table 1:** *Parameter Tuning for Similarity Measures*

### 4.3 Analysis of Classification Accuracy

In order to provide a comprehensive evaluation, we perform the experiments on 38 diverse time series data sets, from the UCR Time Series repository [29], which make up somewhere between 90% and 100% of all publicly available, labelled time series data sets in the world. For several years everyone in the data mining/database community has been invited to contribute data sets to this archive, and 100% of the donated data sets have been archived. This ensures that the collection represents the interest of the data mining/database community, and not just one group. All the data sets have been normalized to have a maximum scale of 1.0 and all the time series are z-normalized. The entire simulation was conducted on a computing cluster at Northwestern University, with 20 multi-core workstations running for over a month. The results are presented in Table 2, such as the standard deviation of the cross validations, are hosted on our web site [1].

To provide a more intuitive illustration of the performance of the similarity measures compared in Table 2, we now use scatter plots to conduct pair-wise comparisons. In a scatter plot, the error ratios of the two similarity measures under comparison are used as the $x$ and $y$ coordinates of a dot, where each dot represents a particular data set. Where a scatter plot has the label "A vs B", a dot above the line indicates that A is more accurate than B (since these are error ratios). The further a dot is from the line, the greater the margin of accuracy improvement. The more dots on one side of the line indicates that the worse one similarity measure is compared to the other .

**Table 2:** Error Ratio of Different Similarity Measures on 1NN Classifier

[a] DTW (c) denotes DTW with constrained warping window, same for LCSS.

| Data Set | crosses# | ED | L$_1$-norm | L$_\infty$-norm | DISSIM | TQuEST | DTW | DTW (c)[a] | EDR | ERP | LCSS | LCSS(c) | Swale | Spade |
|---|---|---|---|---|---|---|---|---|---|---|---|---|---|---|
| 50words | 5 | 0.407 | 0.379 | 0.555 | 0.378 | 0.526 | 0.375 | 0.291 | 0.271 | 0.341 | 0.298 | 0.279 | 0.281 | 0.341 |
| Adiac | 5 | 0.464 | 0.495 | 0.428 | 0.497 | 0.718 | 0.465 | 0.446 | 0.457 | 0.436 | 0.434 | 0.418 | 0.408 | 0.438 |
| Beef | 2 | 0.4 | 0.55 | 0.583 | 0.55 | 0.683 | 0.433 | 0.433 | 0.4 | 0.567 | 0.402 | 0.517 | 0.384 | 0.5 |
| Car | 2 | 0.275 | 0.3 | 0.3 | 0.217 | 0.267 | 0.333 | 0.258 | 0.371 | 0.167 | 0.208 | 0.35 | 0.233 | 0.25 |
| CBF | 16 | 0.087 | 0.041 | 0.534 | 0.049 | 0.171 | 0.003 | 0.006 | 0.013 | 0 | 0.017 | 0.015 | 0.013 | 0.044 |
| chlorineconcentration | 9 | 0.349 | 0.374 | 0.325 | 0.368 | 0.44 | 0.38 | 0.348 | 0.388 | 0.376 | 0.374 | 0.368 | 0.374 | 0.439 |
| cinc_ECG_torso | 30 | 0.051 | 0.044 | 0.18 | 0.046 | 0.084 | 0.165 | 0.006 | 0.011 | 0.145 | 0.057 | 0.023 | 0.057 | 0.148 |
| Coffee | 2 | 0.193 | 0.246 | 0.087 | 0.196 | 0.427 | 0.191 | 0.252 | 0.16 | 0.213 | 0.213 | 0.237 | 0.27 | 0.185 |
| diatomsizereduction | 10 | 0.022 | 0.033 | 0.019 | 0.026 | 0.161 | 0.015 | 0.026 | 0.019 | 0.026 | 0.045 | 0.084 | 0.028 | 0.016 |
| ECG200 | 5 | 0.162 | 0.182 | 0.175 | 0.16 | 0.266 | 0.221 | 0.153 | 0.211 | 0.213 | 0.171 | 0.126 | 0.17 | 0.256 |
| ECGFiveDays | 26 | 0.118 | 0.107 | 0.235 | 0.103 | 0.181 | 0.154 | 0.122 | 0.111 | 0.127 | 0.232 | 0.187 | 0.29 | 0.265 |
| FaceFour | 5 | 0.149 | 0.144 | 0.421 | 0.172 | 0.144 | 0.064 | 0.164 | 0.045 | 0.042 | 0.144 | 0.046 | 0.134 | 0.25 |
| FacesUCR | 11 | 0.225 | 0.192 | 0.401 | 0.205 | 0.289 | 0.06 | 0.079 | 0.05 | 0.028 | 0.046 | 0.046 | 0.03 | 0.315 |
| fish | 5 | 0.319 | 0.293 | 0.314 | 0.311 | 0.496 | 0.329 | 0.261 | 0.107 | 0.216 | 0.067 | 0.16 | 0.171 | 0.15 |
| Gun_Point | 5 | 0.146 | 0.092 | 0.186 | 0.084 | 0.175 | 0.14 | 0.055 | 0.079 | 0.161 | 0.098 | 0.065 | 0.066 | 0.007 |
| Haptics | 5 | 0.619 | 0.634 | 0.632 | 0.64 | 0.669 | 0.622 | 0.593 | 0.466 | 0.601 | 0.631 | 0.58 | 0.581 | 0.736 |
| InlineSkate | 6 | 0.665 | 0.646 | 0.715 | 0.65 | 0.757 | 0.557 | 0.603 | 0.531 | 0.483 | 0.517 | 0.525 | 0.533 | 0.643 |
| ItalyPowerDemand | 8 | 0.04 | 0.047 | 0.044 | 0.043 | 0.089 | 0.067 | 0.055 | 0.075 | 0.05 | 0.1 | 0.076 | 0.082 | 0.233 |
| Lighting2 | 5 | 0.341 | 0.251 | 0.389 | 0.261 | 0.444 | 0.204 | 0.32 | 0.088 | 0.19 | 0.199 | 0.108 | 0.16 | 0.272 |
| Lighting7 | 2 | 0.377 | 0.286 | 0.566 | 0.3 | 0.503 | 0.252 | 0.202 | 0.093 | 0.287 | 0.282 | 0.116 | 0.279 | 0.557 |
| MALLAT | 20 | 0.032 | 0.041 | 0.079 | 0.042 | 0.094 | 0.038 | 0.04 | 0.08 | 0.033 | 0.088 | 0.091 | 0.09 | 0.167 |
| MedicalImages | 5 | 0.319 | 0.322 | 0.36 | 0.329 | 0.451 | 0.286 | 0.281 | 0.36 | 0.309 | 0.349 | 0.357 | 0.348 | 0.434 |
| Motes | 24 | 0.11 | 0.082 | 0.24 | 0.08 | 0.211 | 0.09 | 0.118 | 0.095 | 0.106 | 0.064 | 0.077 | 0.073 | 0.103 |
| OliveOil | 2 | 0.15 | 0.236 | 0.167 | 0.216 | 0.298 | 0.1 | 0.118 | 0.062 | 0.132 | 0.135 | 0.055 | 0.097 | 0.207 |
| OSULeaf | 5 | 0.448 | 0.488 | 0.52 | 0.474 | 0.571 | 0.401 | 0.424 | 0.115 | 0.365 | 0.359 | 0.281 | 0.403 | 0.212 |
| plane | 6 | 0.051 | 0.037 | 0.033 | 0.042 | 0.038 | 0.001 | 0.032 | 0.001 | 0.01 | 0.016 | 0.062 | 0.023 | 0.006 |
| SonyAIBORobotSurface | 16 | 0.081 | 0.076 | 0.106 | 0.088 | 0.135 | 0.077 | 0.074 | 0.084 | 0.07 | 0.228 | 0.155 | 0.205 | 0.195 |
| SonyAIBORobotSurfaceII | 12 | 0.094 | 0.084 | 0.135 | 0.071 | 0.186 | 0.08 | 0.083 | 0.092 | 0.062 | 0.238 | 0.089 | 0.281 | 0.322 |
| StarLightCurves | 9 | 0.142 | 0.143 | 0.151 | 0.142 | 0.13 | 0.089 | 0.086 | 0.107 | 0.125 | 0.118 | 0.124 | 0.12 | 0.142 |
| SwedishLeaf | 5 | 0.295 | 0.286 | 0.357 | 0.299 | 0.347 | 0.256 | 0.221 | 0.145 | 0.164 | 0.147 | 0.148 | 0.14 | 0.254 |
| Symbols | 30 | 0.088 | 0.098 | 0.152 | 0.093 | 0.078 | 0.049 | 0.096 | 0.02 | 0.059 | 0.053 | 0.055 | 0.058 | 0.018 |
| synthetic_control | 5 | 0.142 | 0.146 | 0.227 | 0.158 | 0.64 | 0.019 | 0.014 | 0.118 | 0.035 | 0.06 | 0.075 | 0.06 | 0.15 |
| Trace | 5 | 0.368 | 0.279 | 0.445 | 0.286 | 0.158 | 0.016 | 0.075 | 0.15 | 0.084 | 0.118 | 0.142 | 0.108 | 0 |
| TwoLeadECG | 25 | 0.129 | 0.154 | 0.151 | 0.163 | 0.266 | 0.033 | 0.07 | 0.065 | 0.071 | 0.146 | 0.154 | 0.149 | 0.017 |
| Two-Patterns | 5 | 0.095 | 0.039 | 0.797 | 0.036 | 0.747 | 0 | 0 | 0.001 | 0.01 | 0 | 0 | 0 | 0.052 |
| wafer | 7 | 0.005 | 0.004 | 0.021 | 0.005 | 0.014 | 0.015 | 0.005 | 0.002 | 0.006 | 0.004 | 0.004 | 0.004 | 0.018 |
| WordsSynonyms | 5 | 0.393 | 0.374 | 0.53 | 0.375 | 0.529 | 0.371 | 0.315 | 0.295 | 0.346 | 0.294 | 0.28 | 0.274 | 0.322 |
| yoga | 11 | 0.16 | 0.161 | 0.181 | 0.167 | 0.216 | 0.151 | 0.151 | 0.112 | 0.133 | 0.109 | 0.134 | 0.43 | 0.13 |

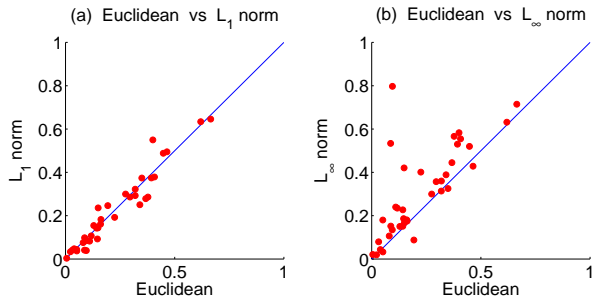

**Fig. 11:** Accuracy of Various $L_p$-norms, above the line Euclidean outperforms $L_1$ norm/$L_\infty$ norm.

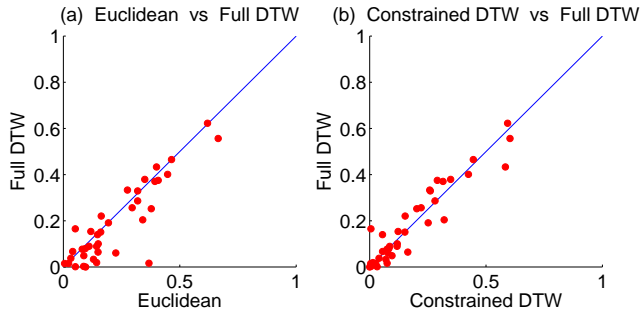

**Fig. 12:** Accuracy of DTW, above the line Euclidean/Constrained DTW outperforms Full DTW.

First, we compare the different variances of $L_p$-norms.

Figure 11 shows that the Euclidean distance and the Manhattan distance have a very close performance, while both largely outperform the $L_\infty$-norm. This is expected, as a consequence of its definition: the $L_\infty$-norm uses the maximum distance between two sets of time series points, and is more sensitive to noise [24].

Next we illustrate the performance of DTW against Euclidean. Figure 12 (a) shows that full DTW is clearly superior over Euclidean on the data sets we tested. Figure 12 (b) shows that the effectiveness of constrained DTW is the same (or even slightly better) than that of full DTW. This means that we could generally use the constrained DTW instead of DTW to reduce the time for computing the distance and to utilize proposed lower bounding techniques [35].

Unless otherwise stated, in the following we compare the rest of the similarity measures against Euclidean distance and full DTW, since Euclidean distance is the fastest and most straightforward measure, and DTW is the oldest elastic measure.

The performance of DISSIM against that of Euclidean and DTW is shown in Figure 13. It can be observed that the accuracy of DISSIM is slightly better than Euclidean distance; however, it is apparently inferior to DTW.

The performance of TQuEST against that of Euclidean and DTW is shown in Figure 14. On most of the data sets, TQuEST is worse than Euclidean and DTW distances. While the outcome of this experiment cannot account for the usefulness of TQuEST, it indicates that there is a need to investigate the characteristics of the data set for which TQuEST is a favorable measure.

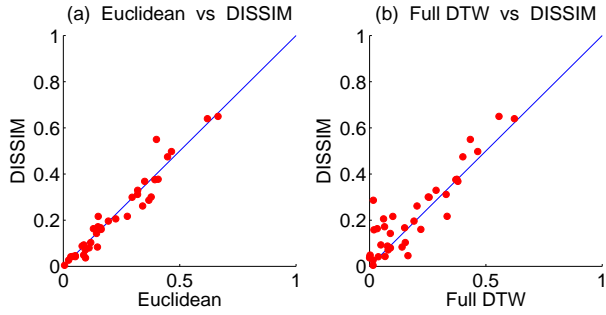

**Fig. 13:** Accuracy of DISSIM, above the line Euclidean/Full DTW outperforms DISSIM.

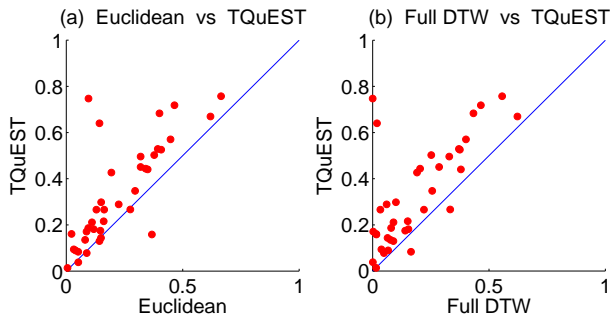

**Fig. 14:** Accuracy of TQuEST, above the line Euclidean/Full DTW outperforms TQuEST.

The respective performances of LCSS, EDR and ERP against the Euclidean and DTW measures are illustrated in Figure 15, Figure 16 and Figure 17, where the left portions of each Figure represent the comparison against the Euclidian distance, and the right portions represent the comparison against DTW. An obvious conclusion is that all three distances outperform the Euclidean distance by a large percentage. However, while it is commonly believed that these edit distance based similarity measures are superior to DTW [13, 15, 17], our experiments, to say the least, demonstrate that this need not be the case in general. As shown, only EDR is *potentially* slightly better than full DTW, whereas the performance of LCSS and ERP are very close to DTW. Even for EDR, a more formal analysis using a two-tailed, paired t-test is required to reach any statistically significant conclusion [57]. We also studied the performance of constrained LCSS, as shown in Figure 18. It can be observed that the constrained version of LCSS is even slightly better than the unconstrained one, while it also reduces the computation cost and gives rise to an efficient lower-bounding measure [62].

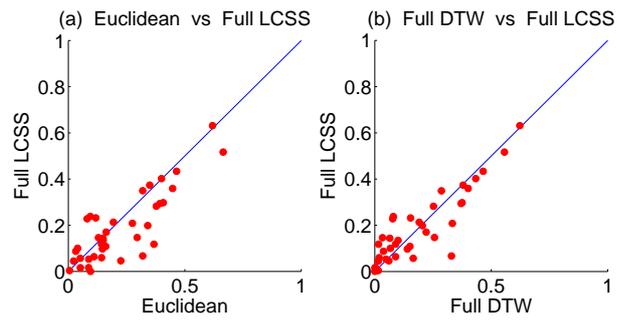

**Fig. 15:** Accuracy of LCSS, above the line Euclidean/Full DTW outperforms Full LCSS.

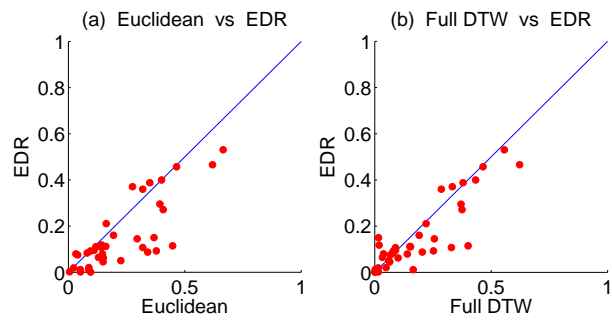

**Fig. 16:** Accuracy of EDR, above the line Euclidean/Full DTW outperforms EDR.

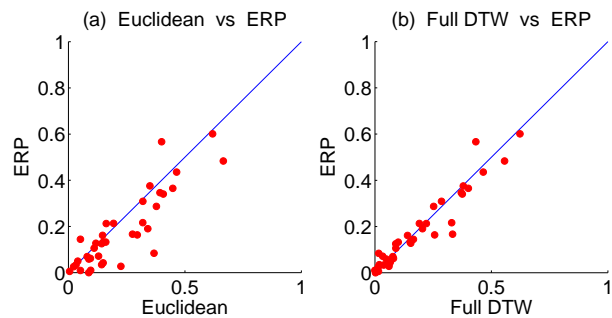

**Fig. 17:** Accuracy of ERP, above the line Euclidean/Full DTW outperforms ERP.

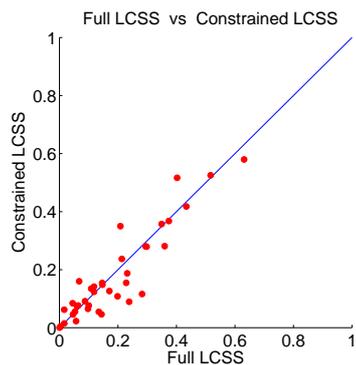

**Fig. 18:** Accuracy of Constrained LCSS, above the line Full LCSS outperforms Constrained LCSS.

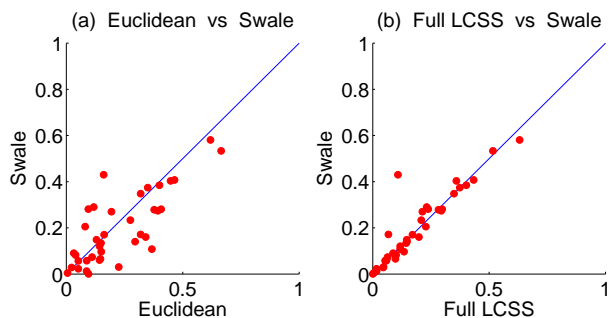

**Fig. 19:** Accuracy of Swale, above the line Euclidean/Full LCSS outperforms Swale.

Next, we compare the performance of Swale against that of Euclidean distance and LCSS, as Swale aims at improving the effectiveness of LCSS and EDR. The results are shown in Figure 19, and suggest that Swale is clearly superior to Euclidean distance, and yields an almost identical accuracy as LCSS.

Finally, we compare the performance of SpADe against that of Euclidean distance and DTW. The results are shown in Figure 20. In general, the accuracy of SpADe is close to that of Euclidean but is inferior to DTW distance, although on some data sets SpADe outperforms the other two. We believe that one of the biggest challenges for SpADe is that it has a large number of parameters that need to be tuned. Given the small tuning data sets, it is very difficult to pick the right values. However, we note again that the outcome of this experiment cannot account for the utility of SpADe. For example, one major contribution of SpADe is to detect interesting patterns online for stream data.

In summary, we found through experiments that there is no clear evidence that one similarity measure exists that is superior to others in the literature in terms of accuracy. While some similarity measures are more effective on certain data sets, they usually inferior on some other data sets. This does not mean that the time series community should settle for the existing similarity measures – quite the contrary. However, we believe that more caution needs to be exercised in order to avoid the possibility of making certain mistakes and drawing invalid conclusions, some of which we address in detail in the subsequent Section 5.

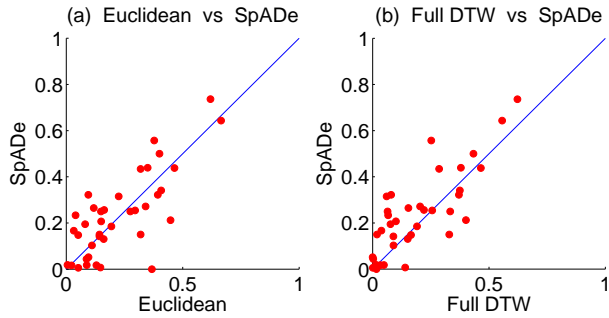

**Fig. 20:** Accuracy of SpADe, above the line Euclidean/Full DTW outperforms SpADe.

## 5 Exploding the Myths of Surrounding Dynamic Time Warping

DTW is one of the earliest similarity measures for time series proposed in the literature. Having shown that, on average, the constrained DTW is no worse than the more recently introduced similarity measures in terms of accuracy across a wide range of problems, we now address some persistent myths about it, including some that have limited its adoption.

### 5.1 DTW is too slow to be of Practical Use

In the literature, it is often claimed that DTW is too slow to be of practical use. Consider the following quotes:

*"...too slow for practical applications, especially when the underlying similarity measure is based on DTW "* [4]. *"The expensive DTW method prohibits high performance and real-time applications"* [44]. *"However, the computational load of DTW is so expensive that it is intractable for many real-world problems"* [23]. *"DTW (is) very expensive, and are not applicable for multi-media data"* [8]. *"computing cost of DTW algorithm is high"* [67], *"DTW-based techniques suffer for performance inefficiencies"* [48].

The literature is replete with similar claims, however, in every case where a particular claim is backed up with an experimental verification, we find (some parts of) the presented evidence unconvincing, to say the least. Consider [49] for example: they test on stock market data, where they use DTW with a warping window of 3% for queries of length 110. Their database has 620 objects. They claim that a lower bound DTW scan takes 8 seconds, whereas their approach takes 0.74 seconds. This seems like an impressive speedup, but when we reproduced their lower bound scan under identical parameters, we found it took 0.011 seconds, some 727 times faster than their claimed time, and much faster than their improvement.

As another example, consider [23], which claims that on an ECG data set with a time series of length 600 and query length 50, DTW takes 552.1 seconds, whereas their proposed approach takes a "mere" 150.78 seconds. However, our experiments showed that we could do DTW on identical conditions in 0.016 seconds, some $34,506$ times faster! Likewise, the results reported in [8] suggest that the average time to find a nearest neighbor in the *GunX*, *Trace*, and *Leaf* data sets is just under a second; however, using nothing more than a linear scan with a lower bound, we find the average time to be 0.009, some 111 times faster. The use of DTW for searching in a handwriting database has been considered in [40], bemoaning the fact that it takes 650 seconds to search through a mere 6,000 words. When we re-did the experiment we found it took 0.328 seconds, which is almost two thousand times faster! Some fraction of this difference could be attributed to improvements in hardware and more efficient code, however the two to three orders of magnitude speed up we can trivially obtain suggest the literature is unduly pessimistic about DTW's speed.

Recall that all our improvements to the pessimistic numbers above come from simply doing a linear scan with a lower bound [30]. The lower bound has been known since 2002, and takes a single line of code [30]. If we indexed the data, the results would be even more dramatic.

The myth of *"DTW is too slow"* seems to come from reading old literature, perhaps combined with implementation bias [34], and it perpetuates itself from paper to paper. However, the fact that we could easily do the many hundreds of millions of DTW calculations required for this paper should help to dispel this myth. Less than one percent of papers in the time series data mining/database literature consider a data set that is larger than $10,000$ objects, yet we can easily search $10,000$ time series of, say, length 256, using DTW – in well under one second.

### 5.2 There is Room to speed up Similarly Search under DTW

Another common and persistent myth about DTW is that it can be further sped up by improving current lower bounds. Needless to say, it is generally the case that tighter lower bounds are better. However, there are diminishing returns for tightness of lower bounds, and as we will show, we have long ago reached a point where it is worthwhile to attempt to improve lower bounds for DTW. More concretely, we argue that the apparent improvements shown in many recent papers [56, 58, 68] are likely to be spurious.

To eliminate the confounding factors of indexing structures, buffering policies etc, we consider the simplest lower bounding search algorithm, which assumes that all the data is in main memory, illustrated by Algorithm 2 below. The algorithm assumes that $C_i$ is the $i^{th}$ time series in database C, which contains $N$ time series, and $Q$ is a query issued to it.

It is easy to see that the time taken for this search algorithm depends only on the data itself, and the tightness of the lower bound. If the lower bound is trivially loose, say we hard-code it to zero, then the test in line 4 will always be true, and we must do the expensive DTW calculation in line 5 for every object in the database. In contrast, if the lower bound is relativity tight, then a large fraction of tests in line 4 will fail, and we can skip that fraction of DTW calculations.

---

**Algorithm 2** Lower_Bounding_Sequential_Scan($Q$,C)

---
1: $best\_so\_far = Inf$
2: **for** all sequences in database C **do**
3:     $LB\_dist = lower\_bound\_distance(C_i, Q)$ //Cheap lower bound
4:     **if** $LB\_dist < best\_so\_far$ **then**
5:         $true\_dist = DTW(C_i, Q)$ // Expensive DTW
6:         **if** $true\_dist < best\_so\_far$ **then**
7:             $best\_so\_far = true\_dist$
8:             $index\_of\_best\_match = i$

---

There are a handful of ways to slightly speed up this algorithm. For example, both DTW and most of the lower bounds can be implemented as early abandoning [36], and we could first sort the time series objects by their lower bound distance before entering the loop in line 2 (this increases the speed at which the *best_so_far* decreases, making the test in line 4 fail more often). However, these produce only modest speed-ups, since most of the strength of this simple algorithm comes from having a tight lower bound.

As with different representations (cf. Section 3), the tightness of a lower bound can be measured by a simple ratio $T$:

$$T = lower\_bound\_distance(C_i, Q)/DTW(C_i, Q)$$

It is clear that $0 \leq T \leq 1$. Note that $T$ must be measured by a large random sampling. How tight are current lower bounds? Let us start by considering the envelope-based lower bound (LB_Keogh) introduced in 2002. Its value depends somewhat on $\delta$, the temporal constraint(cf. Section 2.2) and on the data itself. In general, smooth data sets tend to allow tighter lower bounds than noisy ones. However, values of 0.6 are typical.

We are now in a position to consider the question, can DTW search be further sped up by improving the current lower bound, as frequently claimed? Rather than implementing these various bounds and risking criticism of a poor implementation, we perform the following idealized experiment. We imagine that we have a sequence of progressively better lower bounds. In particular, each time we calculate LB_Keogh we also calculate "magic" lower bounds which are tighter. To calculate these tighter lower bounds we must "cheat". We also calculate the true DTW distance and the difference $d$.

$$d = DTW(C_i, Q) - LB\_Keogh(C_i, Q)$$

We can then add in a fraction of the difference to LB_Keogh to see what effect a tighter lower bound will have. Concretely, we create two idealized lower bounds:

$$LB\_Magic\_50 = LB\_Keogh(C_i, Q) + (0.50 \times d)$$
$$LB\_Magic\_100 = LB\_Keogh(C_i, Q) + (1 \times d)$$

Note the following: although the magic lower bounds have been given extra information, they have not been penalized for it in terms of time complexity. They are assumed to take *exactly* as long to compute as LB_Keogh. Furthermore, note what an extraordinary advantage has been given to these lower bounds – LB_Magic_100 is a logically *perfect* lower bound; it cannot be improved upon. Without doing any experiments it is possible to predict the performance of LB_Magic_100. It will only have to do the expensive calculation in line 5 of Algorithm 2 for $O(log_2(N))$ times.

We can test the utility of these lower bounds by searching increasingly large data sets. We measure the wall clock time to find the nearest neighbor to a randomly chosen query (which did not come from the data set), averaging over 30 queries. We used a data set of star light curves [36]. Figure 21 shows the results.

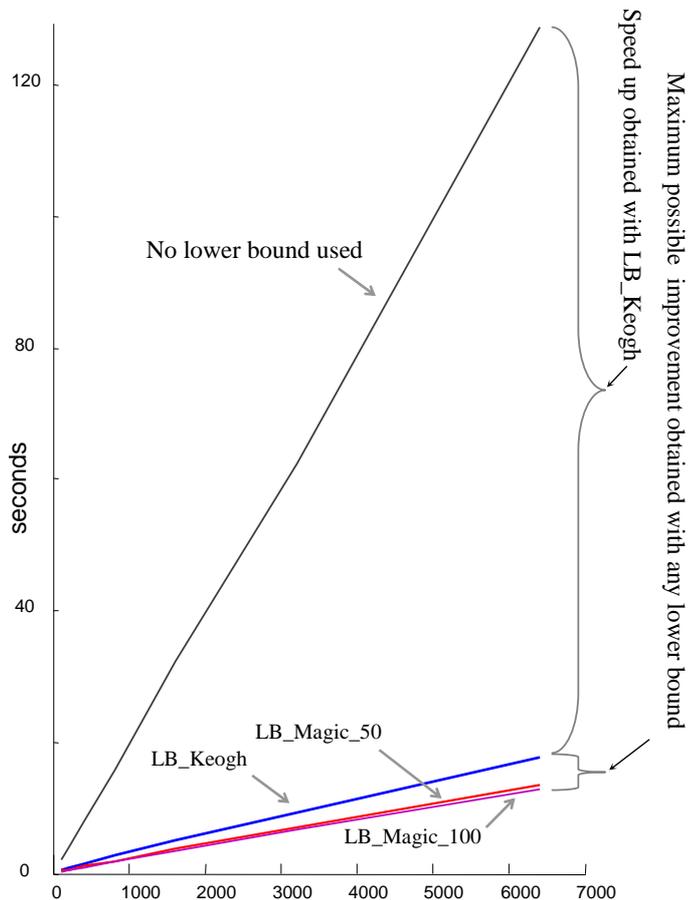

**Fig. 21:** The wall clock time to answer a one-nearest neighbor query in increasingly large instances of a star light curve dataset, for four rival methods[2].

---

[2] In this experiment only the relative times matter. However, the reader may wonder why the absolute times are large. The original time series, which are based on a few dozen (unevenly spaced in time) samples, are greatly over sampled to a length 1,024 by the astronomers as a side effect of their interpolation/smoothing algorithm, and we used a pessimistic temporal constraint of $\delta = 10\%$. If we re-sample them to a more reasonable length of 256, and use the

.

As we can see, the first (non-trival) lower bound introduced for DTW in 2002 really does produce a significant speedup. However, even if we use the idealized optimal lower bound, the *most* of an improvement we could ever hope to obtain is a search that is 1.37 times faster. These results are hard to reconcile with some claims in the literature. For example, [56] claims *"FTW is significantly faster than the best existing method, up to 222 times"*, paper [69] claims improvements of a *"factor of 25"* and paper [42] suggests a more modest improvement of a factor of ten.

However, the results in Figure 21 tell us that such speedups are simply not possible, based only on improving the lower bounds. It might be argued that this result is an anomaly of some kind; however, essentially the same results are seen on other data sets. We see the same basic pattern regardless of the data set, the value of the temporal constraint, the length of the time series, the size of the data set, etc. To our knowledge, there is only one paper that offers a plausible speedup based on a tighter lower bound – [41] suggests a mean speedup of about 1.4 based on a tighter bound. These results are reproducible, and testing on more general data sets we obtained similar results (speedups of between 1.0 and 1.3).

We note that other independent research with scrupulously fair and reproducible findings have confirmed these claims. For example, [6] discovered that while the boundary-based lower bounds introduced in [68] offer slightly tighter bounds, no speed-up could be obtained due to the overhead in the slightly more expensive lower bound calculations. Likewise, while the FTW lower bounds introduced in [56] may offer tighter bounds, similarity search under FTW is significantly slower (about an order of magnitude) than using just LB_Keogh. Quite simply, the large overhead in creating the lower bound here does not manage to break even. In fact, for reasonable values of the threshold $\delta$, the FTW lower bound can take longer to calculate than the original DTW distance!

In summary, while it may be possible to speed up similarity search for DTW (see [6] for example), it is *not* possible to do so significantly by tightening the lower bounds. All claims to the contrary in the literature, to say the least, need to be seriously re-evaluated.

### 5.3 There is Room to produce a more Accurate Distance Measure than DTW?

Our comparative experiments have shown that while elastic measures are, in general, better than lock-step measures, there is little difference between the various elastic measures. This result explicitly contradicts many papers that claim to have a distance measure that is better than DTW, the original and simplest elastic measure. How are we to reconcile these two conflicting claims? We believe the following demonstration will shed some light on the issue. We classified 20 of the data sets hosted at the UCR archive, using the suggested two-fold splits that were establish several years ago. We used a distance measure called ANA (explained below) which has a single parameter that we adjusted to get the best performance. Figure 22 compares the results of our algorithm with Euclidean distance.

As we can see, the ANA algorithm is consistently better than Euclidean distance, often significantly so. Furthermore, ANA is as fast as Euclidean distance, is indexable

---

(empirically) best temporal constraint of $\delta = 4\%$, we can find the nearest neighbor in a data set of size 6, 400 in well under second. This would also *improve* the accuracy(cf.Table 2) and only has a single parameter. Given this, can a paper on ANA be published in a good conference or journal? It is time to explain how ANA works. We downloaded the mitochondrial DNA of an animal from Genbank (www.ncbi.nlm.nih.gov/genbank/). We converted the DNA to a string of integers, with A (Adenine) = 0, C (Cytosine) = 1, G (Guanine) = 2 and T (Thymine)= 3. So the DNA string $GATCA\ldots$ becomes $2, 0, 3, 1, 0, \ldots$.

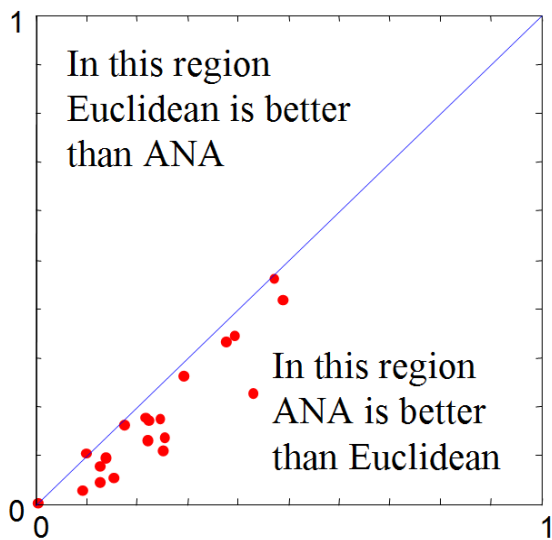

**Fig. 22:** The classification accuracy of ANA compared to Euclidean distance.

Given that we have a string of 16564 integers, we can use the first n integers as weighs when calculating the weights of the Euclidean distance between our time series of length n. So ANA is nothing more than the weighed Euclidean distance, weighed by the DNA string. More concretely, if we have a string S: S = 3, 0, 1, 2, 0, 2, 3, 0, 1, . . . and some time series, say of length 4, then the weight vector W with p = 1 is 3, 0, 1, 2, and the ANA distance is simply:

$$ANA(A, B, W) = \sqrt{\sum_{i=1}^{4}(A_i - B_i)^2 \times W_i}.$$

After we test the algorithm, if we are not satisfied with the result, we simply shift the first location in the string, so that we are using locations 2 to $n + 1$ of the weight string. We continue shifting until the string is exhausted and report the best result in Figure 22. At this point the reader will hopefully say "but that is not fair, you cannot change the parameter after seeing the results, and report the best results". However, we believe this effect may explain many of the apparent improvements over DTW, and in some cases the authors have explicitly acknowledged this [15]. Researchers *are* adjusting the parameters after seeing the results on the test set. In summary, based on the experiments conducted in this paper and all the reproducible fair experiments in the literature, there is no large body of evidence that any distance measure that is systematically better than DTW in general. Furthermore, there is at best very scant evidence that there is any distance measure that is systematically better than DTW in particular domains (say, just ECG data, or just noisy data).

Finally, we are in a position to explain what ANA stands for. It is an acronym for *Arbitrarily Naive Algorithm*. Could these results actually explain the apparently optimistic results in the literature? While we can never know for sure, we can present two pieces of supporting evidence.

**A known example of peeking at the test data:** In a recent paper on time series classification, a new distance measure was introduced, called DPAA [25]. The new measure was tested on the twenty datasets in the UCR archive and it was reported that: "*there are (nine) datasets where DPAA outperforms DTW*". The authors were presented with an earler draft of this chapter, and were asked whether it would be possible that some fraction of their reported results could be attributed to simply adjusting the parameters after seeing the results of the test set. The authors of [25], who were extraordinarily gracious and cooperative (for which we are indebted) did acknowledge that they had adjusted some parameters after looking at the test set, but they did so being confident that this was irrelevant, and their results would hold [26]. They agreed to rerun all the experiments in a strictly blind fashion – they would learn the parameters required by their method only from the training set, and use those parameters for classifying the test data. A week later, after carefully conducting the experiments, they wrote to us, ruefully noting "*according to the new results, DTW (always) outperforms DPAA*" acknowledging that the apparent utility of their method may have resulted due to some omissions in the assumptions which could affect the fairness of the experiments.

**A speculative example of peeking at the test data:** The above example is a rare case where we can be sure that feedback from the test data occurred. We can use this example and an original experiment to ask what feedback from the test data might look like in the more general case. Consider the bottom-right portion of Figure 23. It shows a distance measure compared to constrained DTW on 20 datasets from the UCR archive (essentially a subset of the datasets considered here, but each with one fixed training/test split). If we dismiss the one poor showing (perhaps we could reasonably explain it away as having been an unreliable result on the smallest dataset), then we might take this figure as evidence of the superiority of ATW. After all, it is generally not worse than constrained DTW, and on 3 or 4 datasets it appears to be better. While the visual evidence for ATW is only suggestive, the three companion figures offer *stronger* evidence of superiority. Let us consider them one by one.

- **DTW$_{cheat}$:** Here we compared constrained DTW with itself; however, we "cheated" by adjusting the single parameter requirement *after* seeing the testing results. Even though there is only one (relatively insensitive, see [54]) parameter for us to play with here, the results are apparently better, and without the knowledge of our cheating, a reader might assume that DTW$_{cheat}$ algorithm is a useful contribution.
- **DPAA:** Here we compared constrained DTW with the published results for DPAA discussed in the previous section [25]. Although the results appear to be more of a mixed bag, on a few datasets DPAA does appear to be better. However, recall that when the authors reran the experiments, only looking at the training data, the results were consistently worse than constrained DTW.
- **SpADe:** Here we compared constrained DTW with the results published in [17][3]. In this case the results do appear to offer more hope for optimism. Several of the results really do seem significantly better than constrained DTW. However, for this case we have some additional data. The authors tested this approach on a rigorously fair blind test, the SIGKDD challenge held in 2007 [3]. Here the authors were only given the labeled training data with the *unlabeled* testing data, and had to submit the class predictions for the unlabeled testing data to an independent judge. In this fair test they lost to Euclidean distance on 9 out of the 20 problems, and lost to constrained DTW on 17 out of the 20 problems (the three wins were by the very small margins of 2.0%, 0.4% and 0.1%). Perhaps we could attempt to explain this discrepancy away by assuming that the datasets in the SIGKDD challenge data are very different to the datasets in the UCR archive, and the results in the figure below really *do* represent real cases where the method is better. However, this is not the case. Unknown to all participants in the contest, most of the datasets used in the contest are the same ones that had been publicly available at the UCR archive for many years, just with minor changes in train/test splits to make them less recognizable. For example, in [17] two of the best datasets for their method are Adiac and FACE, where they report significant improvements over Euclidean distance and constrained DTW. However, when faced with a minor rearrangement of these two datasets in the blind contest, the quality of results plunged. The SpADe method got 0.098 error for FACE, whereas Euclidean distance got about half that error (0.0447) and cDTW did even better (0.043). Likewise, for Adiac, a dataset known to be highly "warped", they got 0.3039, which is the same as Euclidean distance, but orders of magnitude worse than constrained DTW, which got just 0.0654 error.

We can now revisit the question asked above; do we think that, at least for some problems, ATW represents an improvement over constrained DTW? We hope that simply the three case studies we have just shown will give the reader pause.

The point of this section is simply to suggest to the community that experimental studies that do not very explicitly state how the parameters were set are of very limited value. Under such circumstances the reader simply cannot decide if the method is making a contribution.

## 6 Conclusion & Future Work

In this paper, we conducted an extensive experimental consolidation on the state-of-the-art representation methods and similarity measures for time series data. We re-implemented and evaluated 8 different dimension-reduction representation methods, as well as 9 different similarity measures and their variants. Our experiments were carried on 38 diverse time series data sets from various application domains. Based on the experimental results we obtained, we make the following conclusions:

1. The tightness of lower bounding, thus the pruning power, thus the indexing effectiveness of the different representation methods for time series data have, for the most part, very little difference on various data sets.

2. For time series classification, as the size of the training set increases, the accuracy of elastic measures converge with that of Euclidean distance. However, on small data sets, elastic measures, e.g., DTW, LCSS, EDR and ERP etc. can be significantly

---

[3] Only 17 of the 20 datasets were available at the time this was published.

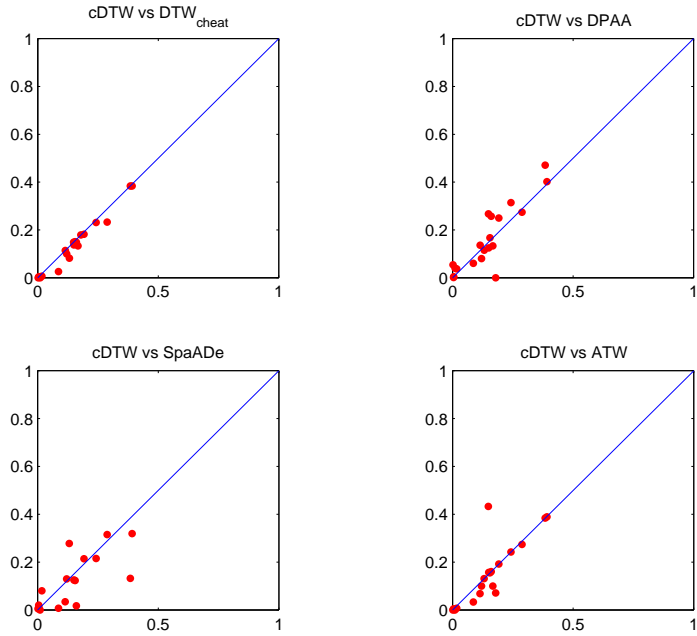

**Fig. 23:** Accuracy of constrained DTW vs. four other approaches, above the line constrained DTW is better, below the line the rival approach wins.

more accurate than Euclidean distance and other lock-step measures, e.g., $L_\infty$-norm, DISSIM.

3. Constraining the warping window size for elastic measures, such as DTW and LCSS, can reduce the computation cost and enable effective lower-bounding, while yielding the same or even better accuracy.

4. The accuracy of edit distance based similarity measures, such as LCSS, EDR and ERP, are *very* close to that of DTW, a 40-year-old, much simpler technique.

5. The accuracy of several novel types of similarity measures, such as TQuEST and SpADe, are in general inferior to elastic measures.

6. If a similarity measure is not accurate enough for the task, getting more training data *really* helps. This is shown in Figure 10 where the error rate of both DTW and Euclidean distance is reduced by more than an order of magnitude when we go from a training set of size 50 to size 2000.

7. If getting more data is not possible, then trying the other measures *might* help; however, extreme care must be taken to avoid overfitting. If we test enough measures on a single train/test split, there is an excellent possibility of finding a measure that improves the accuracy by chance, but will not generalize.

As an additional comment, but not something that can be conclusively validated from our experiments, we would like to bring up an observation which, we hope, may steer some interesting directions of future work. Namely, when pair-wise comparison is done among the methods, in a few instances we have one method that has worse accuracy than the other in the majority of the data sets, but in the ones that it is better, it does so by a large margin. Could this be due to some intrinsic properties of the data set? If so, could it be that those properties may have a critical impact on which distance measure(s) [50] should be applied? We believe that in the near future, the research community will generate some important results along these lines.

As an immediate extension, we plan to conduct more rigorous statistical analysis on the experimental results we obtained. We will also extend our evaluation on the accuracy of the similarity measures to more realistic settings, by allowing missing points in the time series and adding noise to the data. Another extension is to validate the effectiveness of some of the existing techniques in expediting similarity search using the respective distance measures.

**Acknowledgements** We would like to sincerely thank all the donors of the data sets, without which this work would not have been possible. We would also like to thank Lei Chen, Jignesh Patel, Beng Chin Ooi, Yannis Theodoridis and their respective team members for generously providing the source code to their similarity measures as references and for providing timely help on this project. We would like to thank Michalis Vlachos, Themis Palpanas, Chotirat (Ann) Ratanamahatana and Dragomir Yankov for useful comments and discussions. Needless to say, any errors herein are the responsibility of the authors only. We would also like to thank Gokhan Memik and Robert Dick at Northwestern University for providing the computing resources for our experiments.

We have made the best effort to faithfully re-implement all the algorithms, and evaluate them in a fair manner. The purpose of this project is to provide a consolidation of existing works on querying and mining time series, and to serve as a starting point for providing references and benchmarks for future research on time series. We welcome all kinds of comments on our source code and the data sets of this project [1], and suggestions on other experimental evaluations.

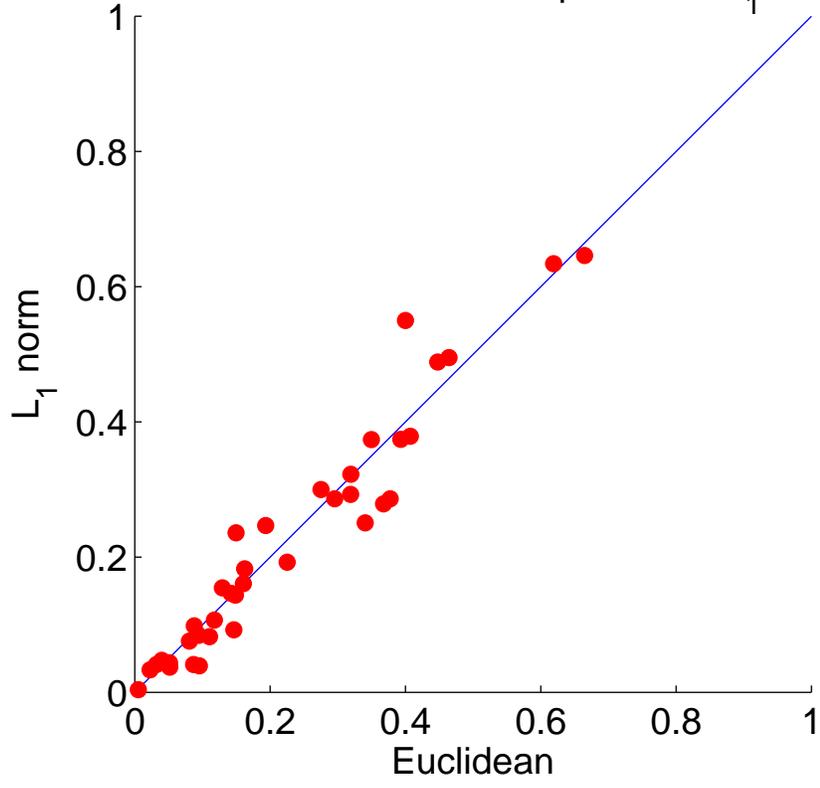

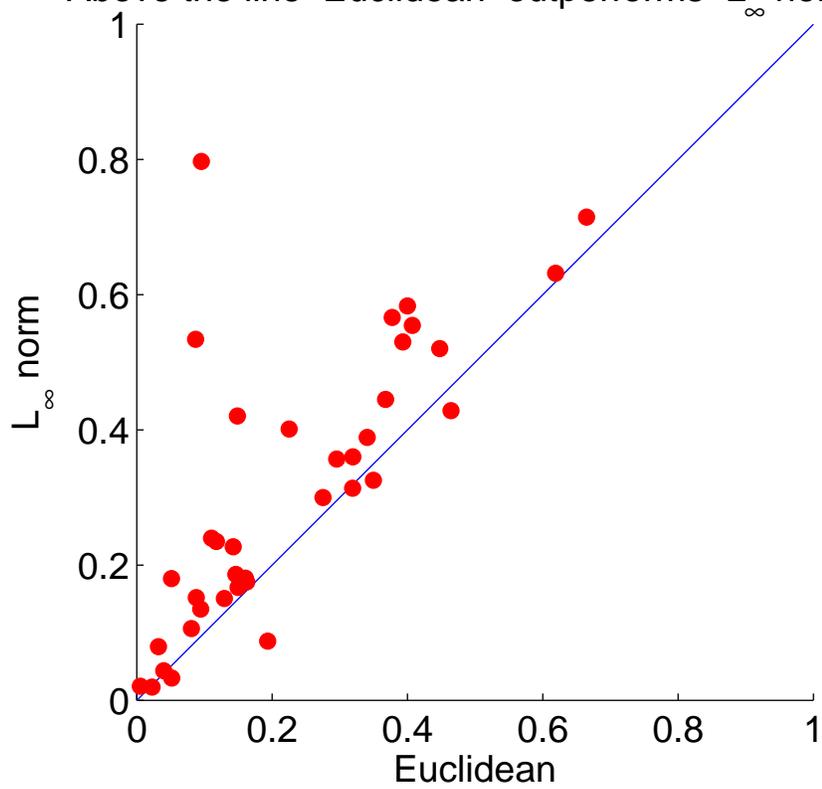

Above the line  Euclidean  outperforms  DISSIM

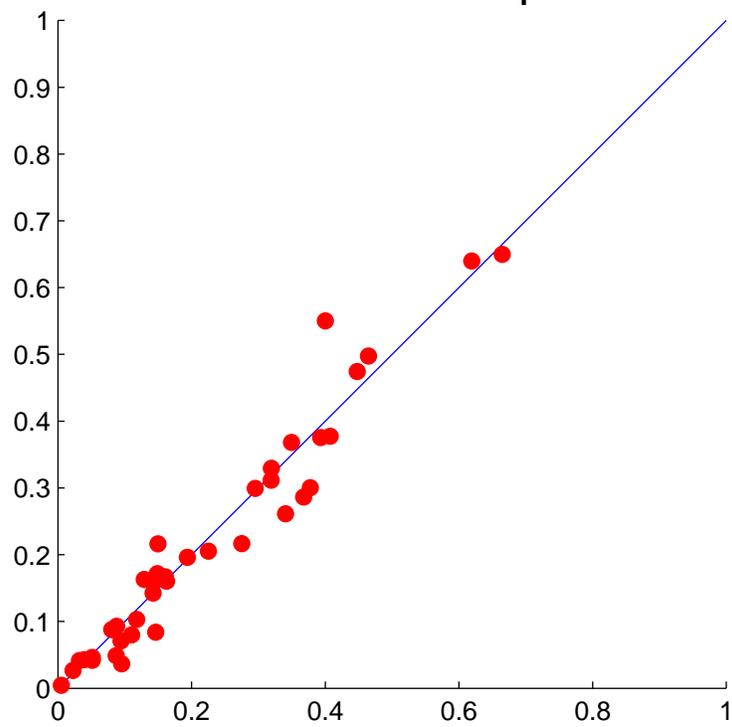

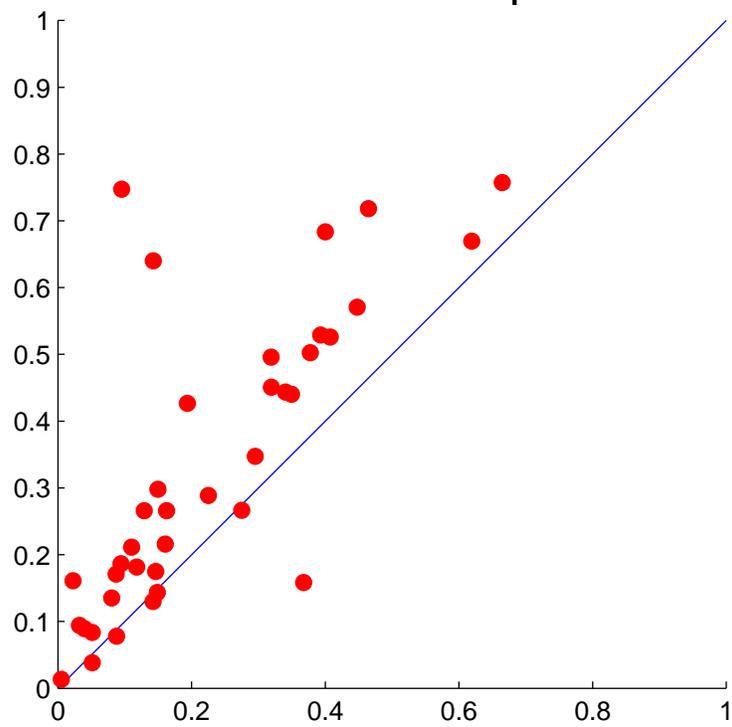

Above the line  Euclidean  outperforms  TQuEST

bove the line  Euclidean  outperforms  constrained D

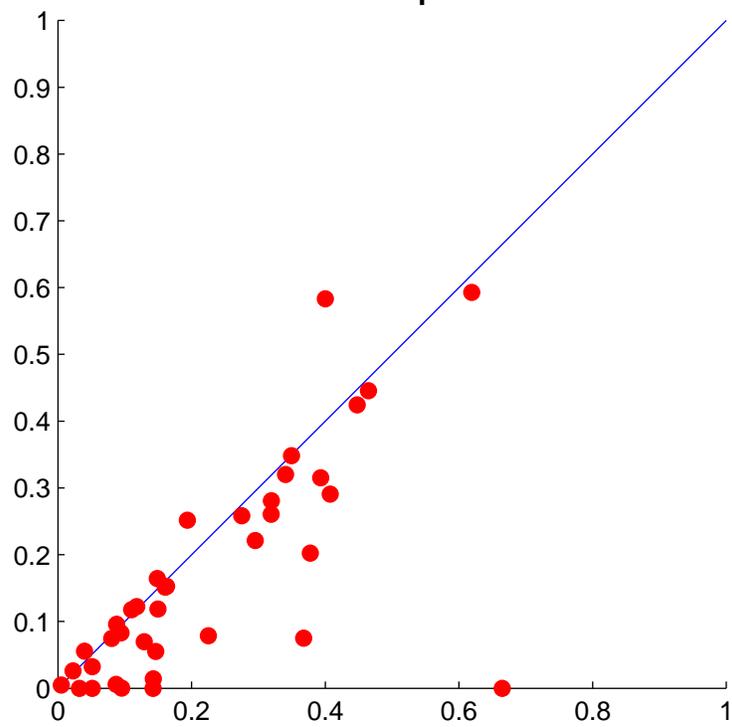

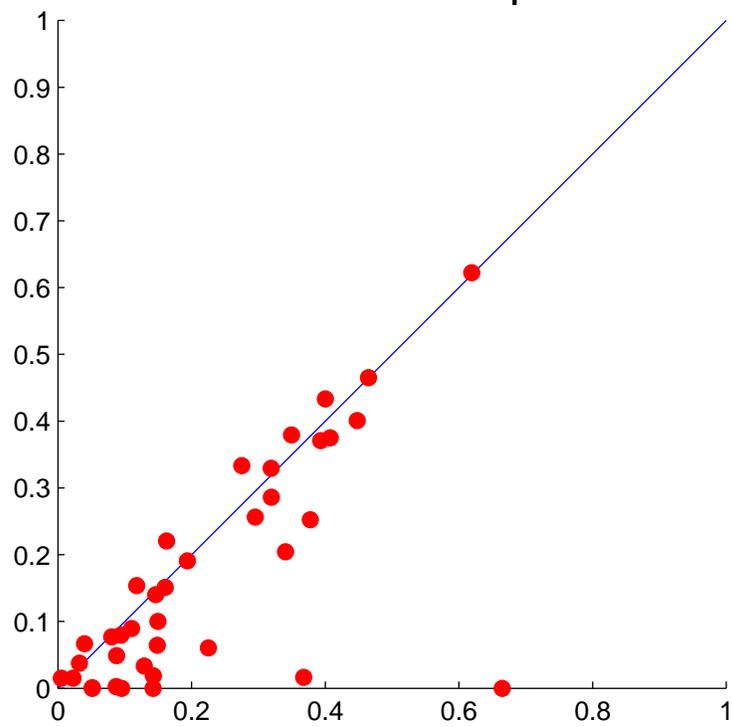

Above the line  Euclidean  outperforms  full DTW

Above the line  DISSIM  outperforms  full DTW

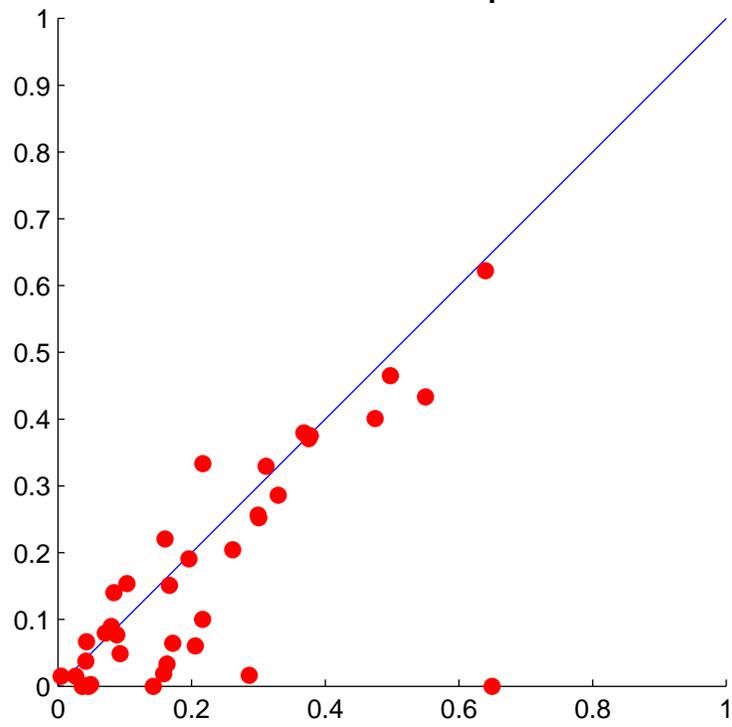

Above the line  TQuEST  outperforms  full DTW

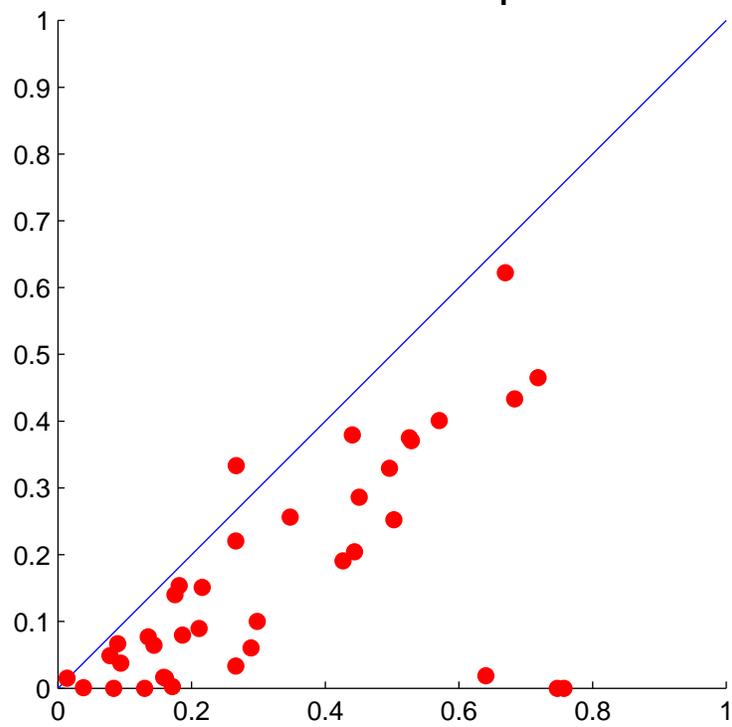

Above the line  constrained DTW  outperforms  full DT
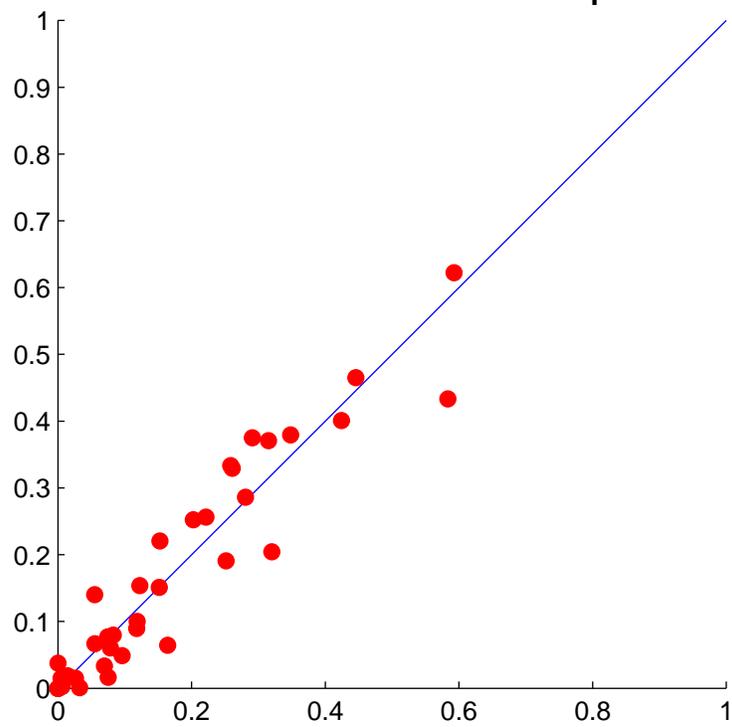

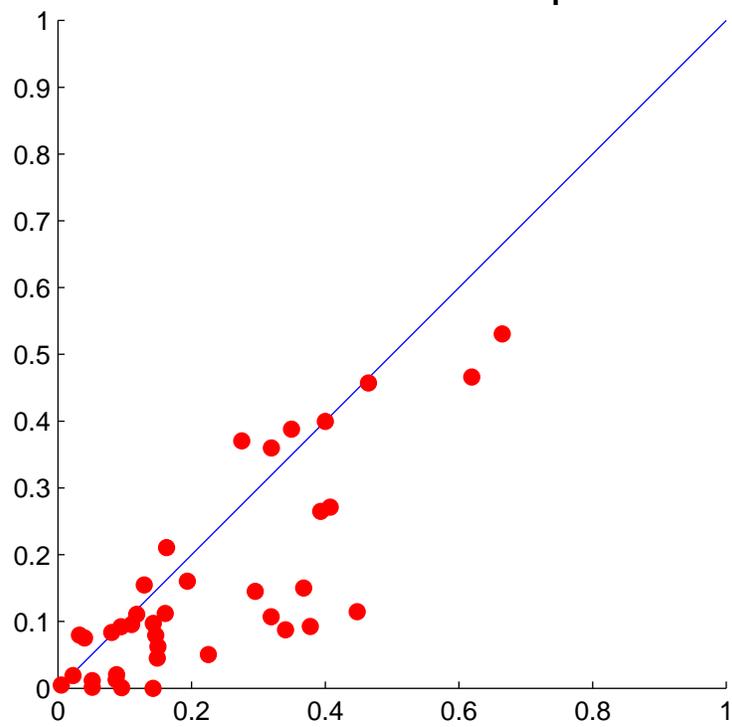

Above the line  full DTW  outperforms  EDR

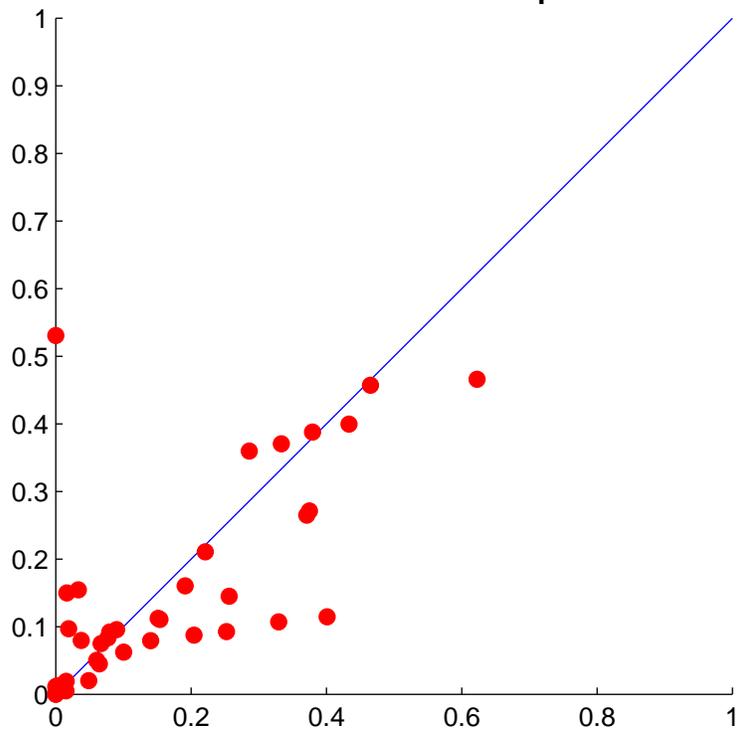

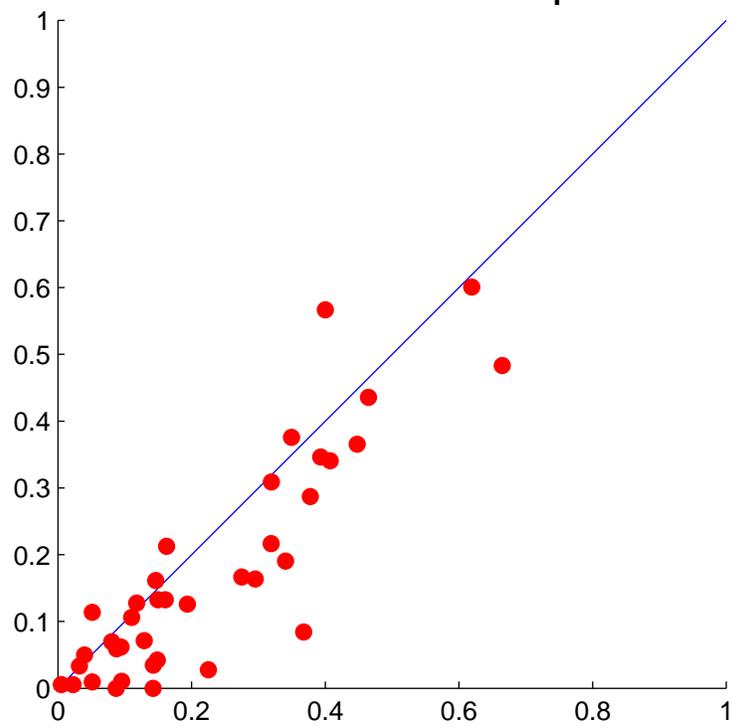
Above the line  Euclidean  outperforms  ERP

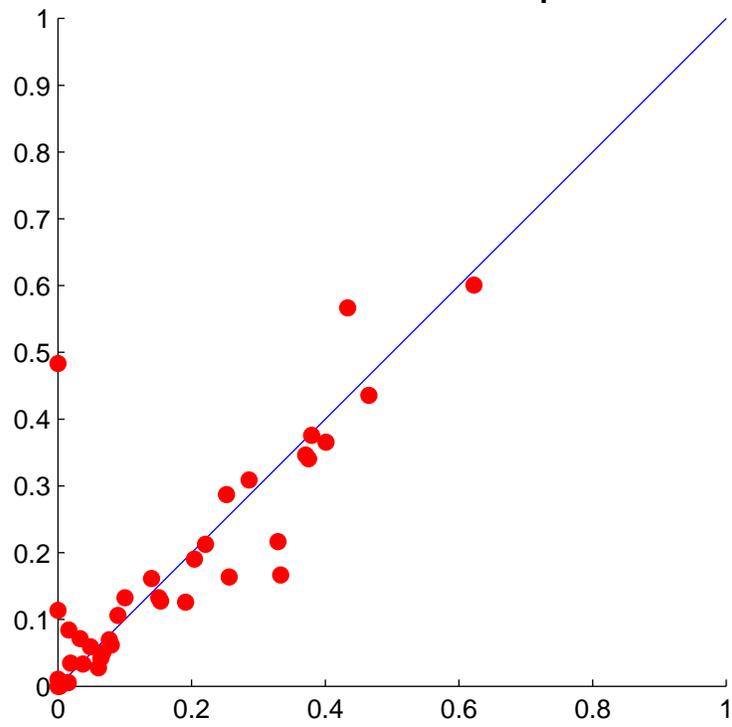

Above the line  full DTW  outperforms  ERP

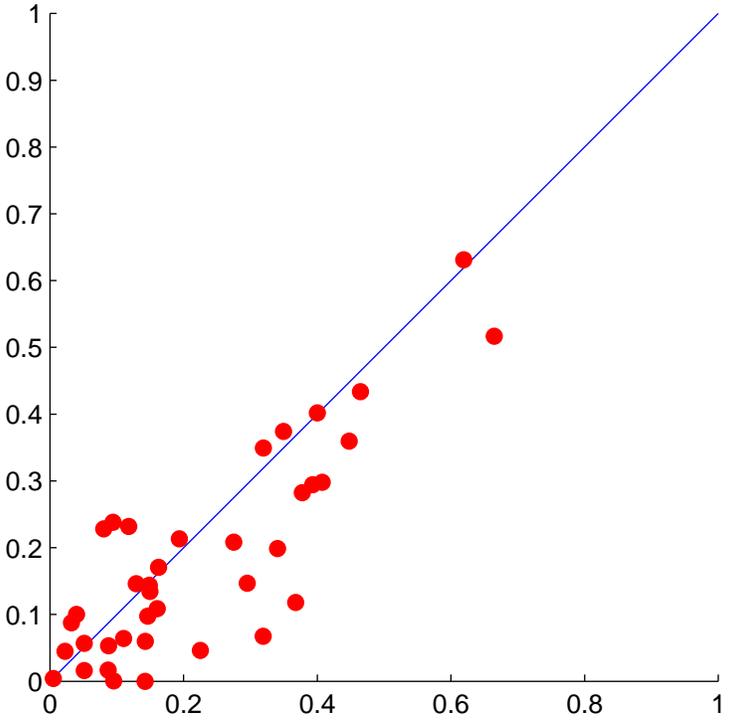

Above the line  full DTW  outperforms  full LCSS

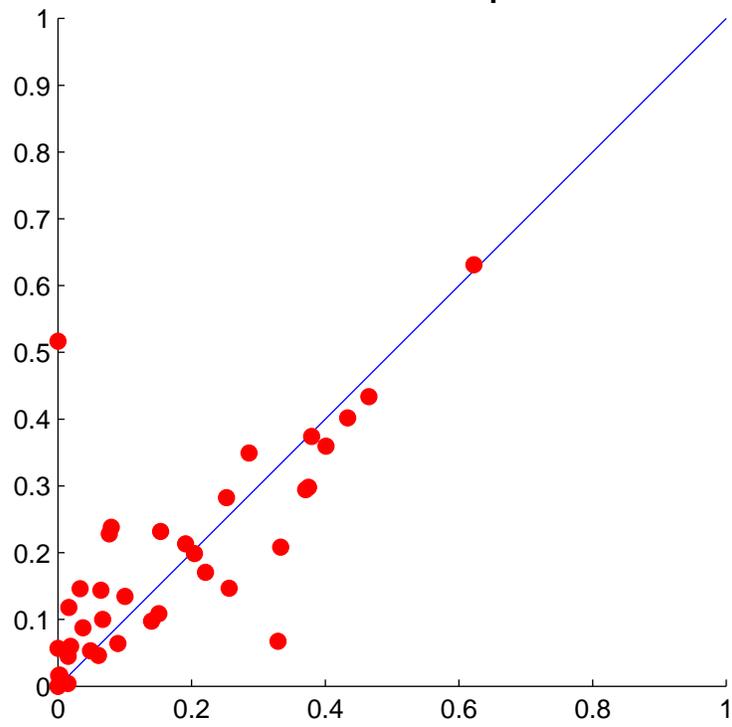

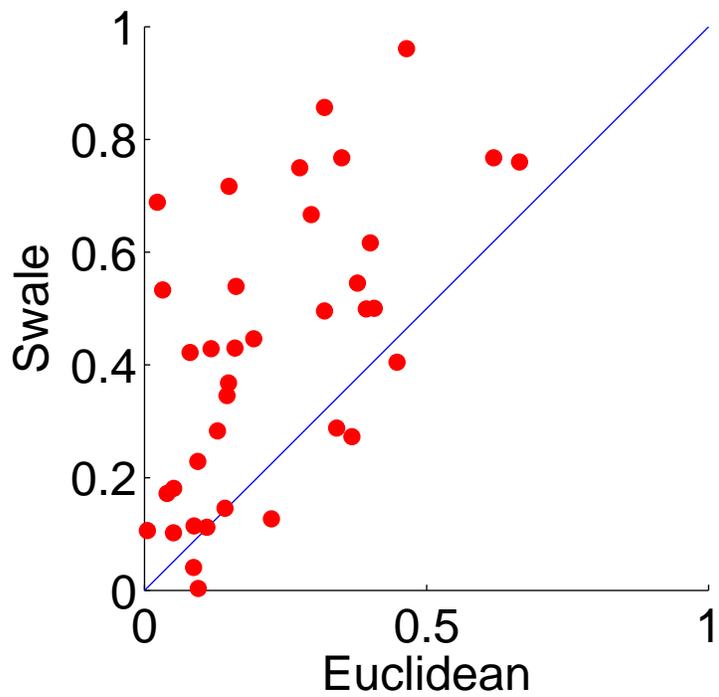 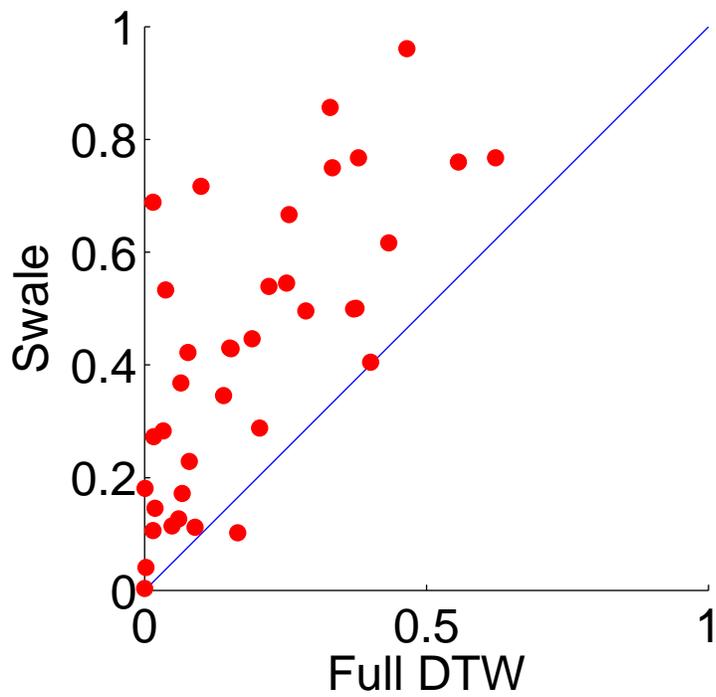

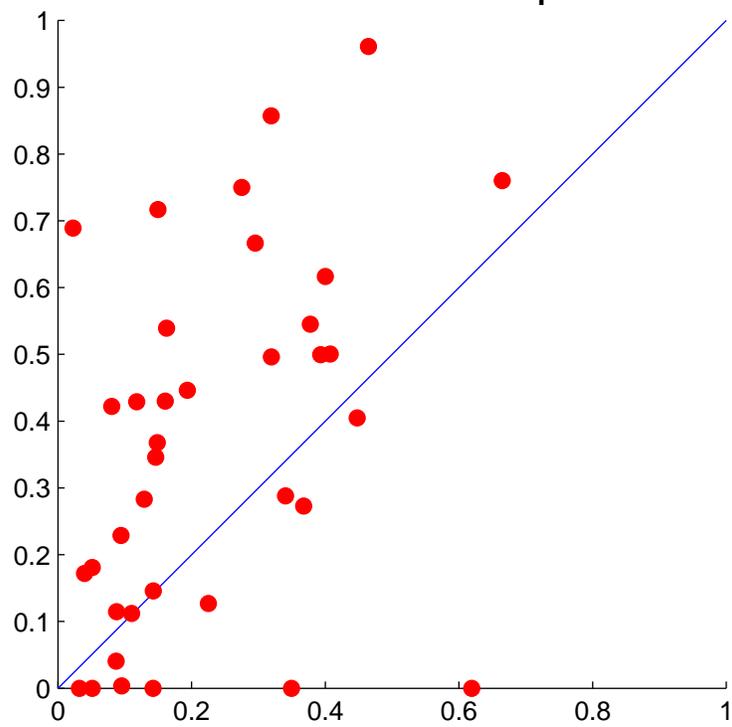
Above the line  Euclidean  outperforms  Swale

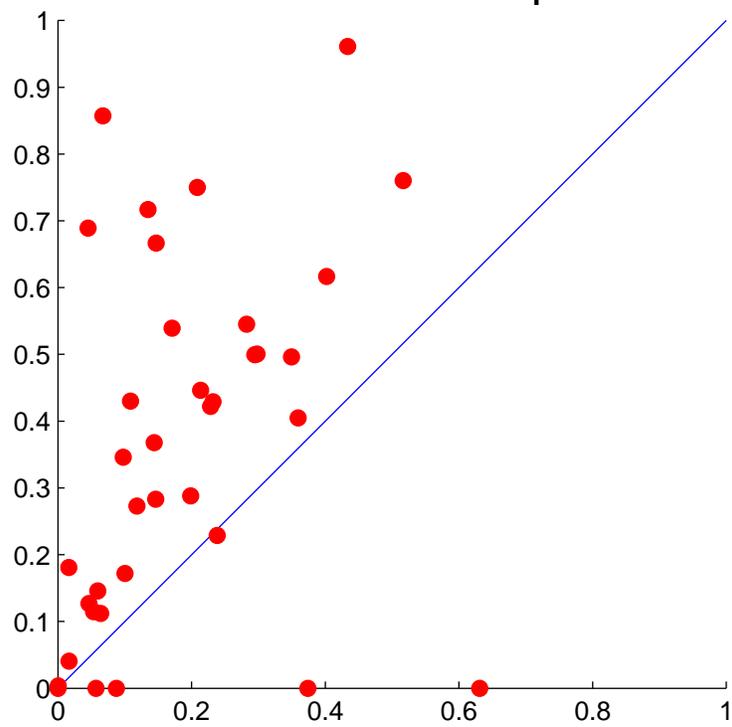

Above the line  full LCSS  outperforms  Swale

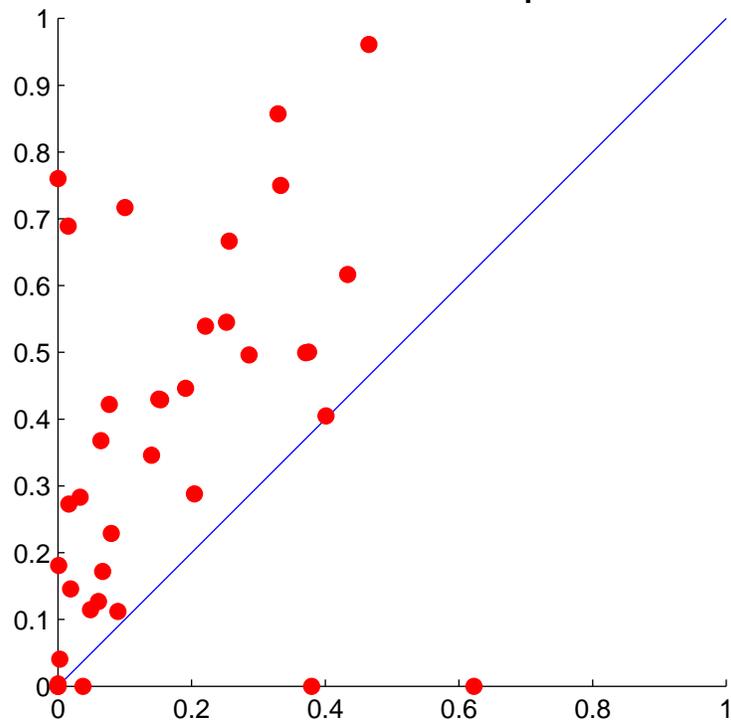

Above the line  full DTW  outperforms  Swale

Above the line  Euclidean  outperforms  SpADe

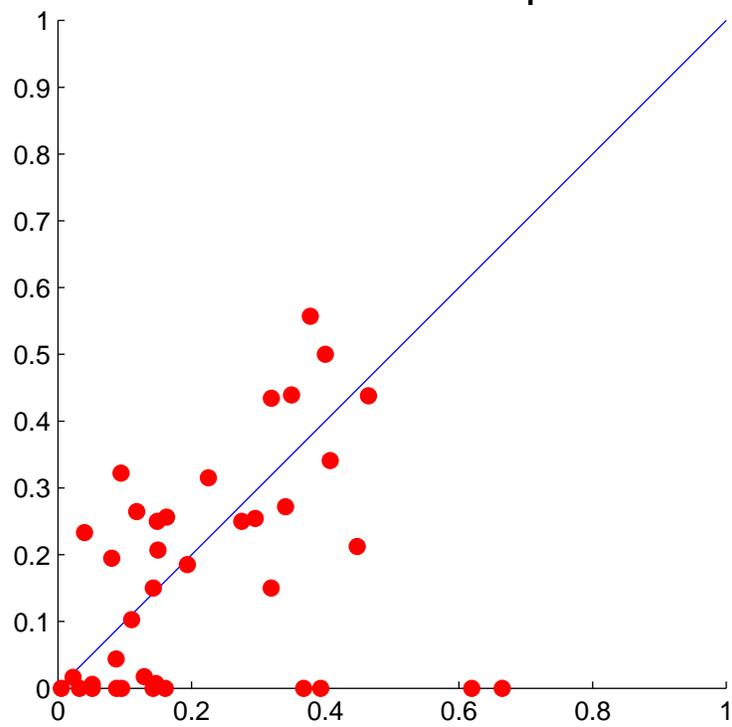

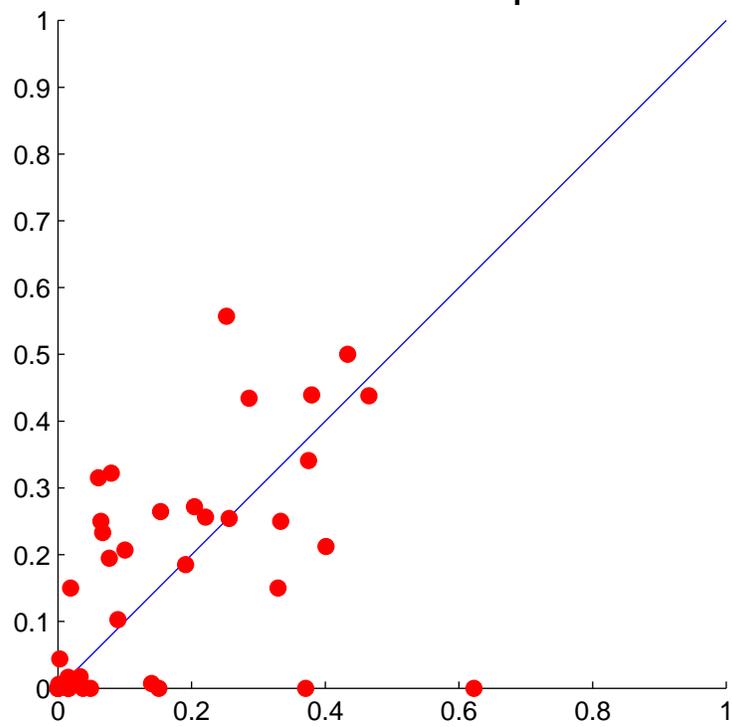

Above the line  full DTW  outperforms  SpADe

In this region
Euclidean is better
than ASA

In this region
ASA is better
than Euclidean

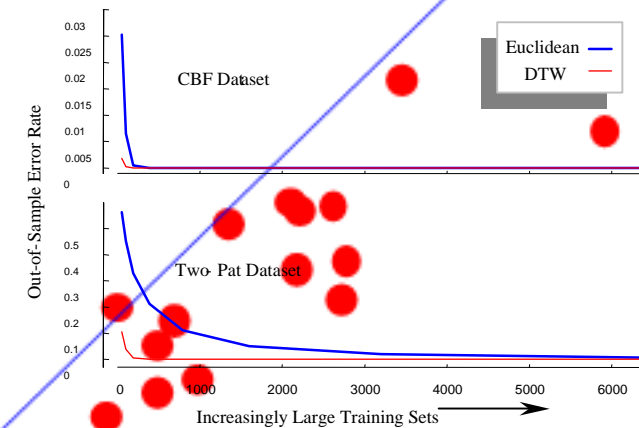

Out-of-Sample Error Rate

CBF Dataset

Euclidean
DTW

Two-Pat Dataset

Increasingly Large Training Sets

0

0                                                                                    1

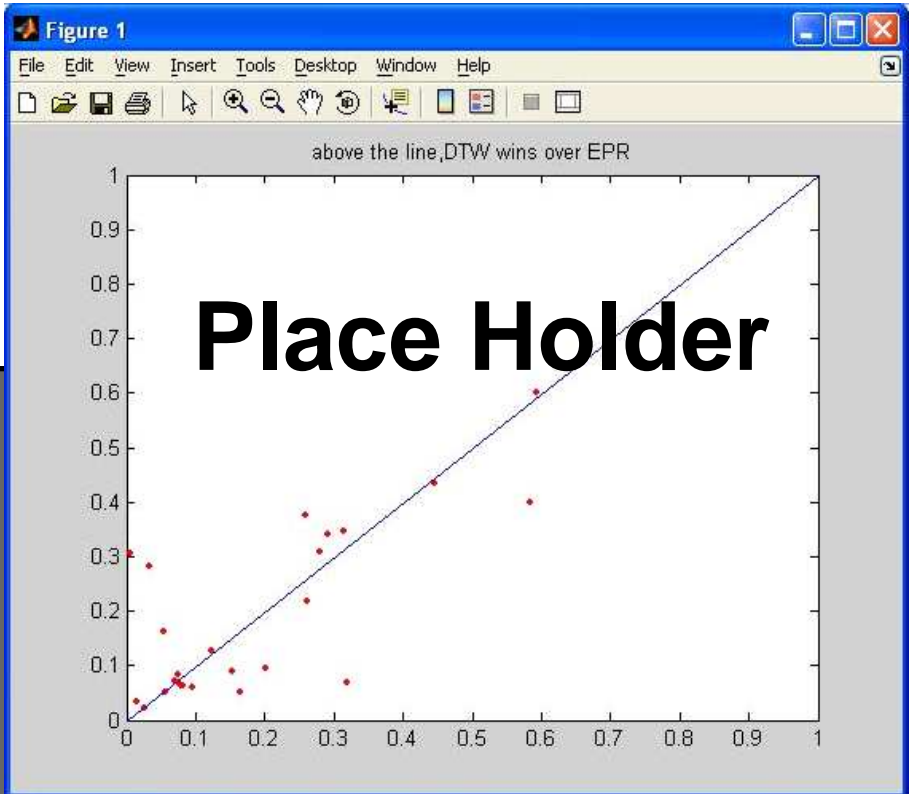